\documentclass[msom,sglanonrev]{informs4}
\RequirePackage{tgtermes}
\RequirePackage{newtxtext}
\RequirePackage{newtxmath}
\RequirePackage{bm}
\RequirePackage{endnotes}

\ifdefined\MILOPAnonymous
\DoubleSpacedXI 
\else
\OneAndAHalfSpacedXII 
\fi

\usepackage{algorithm}
\usepackage{algpseudocode}
\usepackage{tikz}
\usepackage{xcolor}
\ifdefined\MILOPAnonymous
\colorlet{blue}{black} 
\fi
\usepackage{array}
\usepackage{needspace}
\usepackage{cleveref}

\newcommand{\choarev}[1]{{\color{black}#1}}

\usepackage{natbib}
 \bibpunct[, ]{(}{)}{,}{a}{}{,}%
 \def\bibfont{\small}%
\newcommand{\choahead}[2]{\csname #1\endcsname{\choarev{#2}}}

\newcommand{\Ind}[1]{\mathbb{I}\{ {#1} \}}
\newcommand{\E}{\mathbb{E}} 
\usepackage{comment}
\newcommand{\States}{\mathcal{S}}
\newcommand{\Actions}{\mathcal{A}}

\EquationsNumberedThrough    

\TheoremsNumberedThrough     
\ECRepeatTheorems  %

\ifdefined\MILOPAnonymous
\MANUSCRIPTNO{}
\else
\MANUSCRIPTNO{MSOM-0001-2024.00}
\fi

\makeatletter
\def\theARTICLETOP{\vspace*{-12pt}}
\RRHSecondLine{}
\LRHSecondLine{}
\def\theARTICLEABSTRACT{%
  \HOOKb
  \vspace*{18pt}
  \begin{minipage}[t]{\textwidth}\parindent1em
    \ABSfont
    \noindent\theABSTRACT\endgraf
    \vskip5pt
    \theFUNDING
    \theKEYWORDS
    \theSUBJECTCLASS
    \theAREAOFREVIEW
    \theMSCCLASS
    \theORMSCLASS
  \noindent  \hrulefill
  \end{minipage}%
  \vspace*{0pt}
  }
\makeatother

\begin{document}


\ECRUNAUTHOR{}
\ECAUpunct{}

\RUNAUTHOR{Pandey et al.}
\RUNTITLE{Who Gets Watched?}

\TITLE{Who Gets Watched? Allocating Scarce Escalation Resources on Pediatric Wards with Neural Index Policies}

\ARTICLEAUTHORS{%
\AUTHOR{Himadri S. Pandey}
\AFF{H. Milton Stewart School of Industrial and Systems Engineering,
Georgia Institute of Technology, Atlanta, GA, \EMAIL{hpandey30@gatech.edu}}

\AUTHOR{Mark V. Mai}
\AFF{Children's Healthcare of Atlanta, Atlanta, GA, \EMAIL{Mark.Mai@choa.org}}

\AUTHOR{Naveen Muthu}
\AFF{Children's Healthcare of Atlanta, Atlanta, GA}

\AUTHOR{Kai Wang}
\AFF{School of Computational Science and Engineering,
Georgia Institute of Technology, Atlanta, GA, \EMAIL{kwang692@gatech.edu}}

\AUTHOR{Gian-Gabriel P. Garcia}
\AFF{Department of Industrial and Systems Engineering,
University of Washington, Seattle, WA, \EMAIL{garciagg@uw.edu}}
} 
\ABSTRACT{
\textbf{Problem definition:}
Hospitals use separate pathways to identify and respond to deteriorating ward patients. At Children's Healthcare of Atlanta (CHOA), these include Watcher surveillance, Critical Care Outreach Team assessment, and transfer to the intensive care unit. When several patients need attention simultaneously, however, no common process determines who should receive which response under limited staffing and capacity. We study how to coordinate these actions to support earlier escalation without unnecessary monitoring.
\textbf{Methodology/results:}
We model ward escalation decisions using a multi-action restless multi-armed bandit model and develop Multi-action Index Learning and Optimization Policy (MILOP) as a computational solution approach. MILOP trains a neural network to generate a high-quality index policy using a decision-focused learning approach. 
We train and evaluate MILOP using historical CHOA data and compare it with two practical baselines. MILOP recommends an escalation before 87.4\% of critical or emergent intensive-care transfers, compared with 79.5\% for the stronger baseline, while using 13.5\% less monitoring resources. MILOP provides a median of 19.2 hours lead time before transfer, earlier than either baseline. Moreover, 99.3\% of MILOP's proactive transfer recommendations are preceded by Watcher surveillance or outreach-team assessment, demonstrating a desirable escalation of care.
\textbf{Managerial implications:} MILOP can help leaders focus scarce clinical attention, gain more time to respond before severe deterioration, and reduce monitoring burden compared to existing practices. Broadly, hospitals may improve escalation decisions by coordinating existing pathways rather than managing each separately or using risk scores as stand-alone decision rules.  
}





\maketitle


\section{Introduction}
Within pediatric hospital settings, \textit{clinical deterioration}, i.e., a sudden and rapid decline in health, is often considered a significant but preventable source of morbidity and mortality \citep{buist2004association,helen1999unexpected}. 
\choarev{A child whose condition worsens on a hospital ward may need intensive care before the next routine assessment. When deterioration culminates in an urgent transfer, the opportunity for a controlled response may already have narrowed. Pediatric studies associate critical deterioration and emergency transfers with higher in-hospital mortality, motivating earlier recognition and response \citep{bonafide2014critical,mehta2022late}. Early warning systems seek to support that response \citep{lambert2017paediatric}, but their effects depend on what happens after a warning. Specifically, the evidence on the effectiveness of early warning systems is mixed, with several studies including a multicenter randomized trial finding no significant reduction in all-cause hospital mortality \citep{parshuram2018effect,chong2022paediatric}. This evidence emphasizes that, while risk prediction creates an opportunity for higher quality care, realizing improved health outcomes requires an improvement on the decision processes surrounding clinical attention and care escalation.}

We study these decisions in collaboration with Children's Healthcare of Atlanta (CHOA), as part of the Pediatric Technology Center's Pillar 1 program connecting clinical care with data science and artificial intelligence. Specifically, we are co-designing and implementing an algorithm-augmented \textit{watcher process} in which ward teams can designate a child as a \emph{Watcher} for closer surveillance, seek assessment by the critical care outreach team (CCOT), or recommend a proactive transfer to the intensive care unit (ICU). Watcher processes bring clinicians together to identify and review patients of concern \citep{evans2021implementing}, while outreach actions add specialist assessment and support on the ward. The appropriate response depends on both the child's condition and the resources available to act. As such, we aim to provide decision support to facilitate these choices before deterioration occurs and leads to a critical or emergent transfer. Earlier ICU admission can serve that aim even when it does not reduce the total number of transfers. Although these escalation mechanisms exist within CHOA, they have historically operated through separate clinical processes rather than through a single system that jointly coordinates which patients receive which response and when. The watcher process being developed through this partnership creates such a coordinating layer, bringing these distinct resources into a common escalation pathway.

Critically, making these choices within the watcher process go beyond identifying high-risk patients. For example, many early warning scores are focused on estimating the risk of a critical or emergent ICU transfer within 12 hours \citep{lambert2017paediatric} However, ranking patients by that score identifies concern without specifying which action each child should receive especially in a complex watcher process such as the one that we are designing with CHOA. Watcher surveillance, CCOT assessment, and transfer recommendations use different resources and may have different consequences for the patient's subsequent course. Assigning a resource to one child limits the options available to others, and those choices must be revisited as patients improve, deteriorate, arrive, and leave. Because these resources have historically been managed through separate clinical processes, there is no existing rule for jointly deciding how surveillance, outreach, and proactive transfer should be allocated across competing patients. Accordingly, deciding how to convert patient-level risk information into a coordinated, feasible allocation of these limited hospital resources is a key operational challenge for the watcher process being developed at CHOA.

In response to these challenges, we propose an algorithmic framework that integrates an upstream early warning score system with downstream dynamic decision support for CHOA's watcher process. Specifically, we formulate the downstream problem of dynamic patient monitoring and allocation of resources across patients on the ward as a multi-action restless multi-armed bandit (RMAB) model. To generate a high-quality and computationally tractable solution to this RMAB model, we develop a \underline{M}ulti-action \underline{I}ndex \underline{L}earning and \underline{O}timization \underline{P}olicy (MILOP) approach. MILOP learns how to generate an implementable multi-action index policy by combining a neural network with differentiable resource allocation via optimal transport methods. In an empirical analysis using CHOA's historical data, we find that our proposed framework shows significant improvement over competitive benchmarks across several clinically relevant metrics. For example, compared with \textit{threshold-based top-$k$} and \textit{top-$k$} risk ranking, MILOP recommends escalation before an observed critical or emergent ICU transfer in a greater proportion of episodes (87.4\% vs. 79.5\% and 70.2\%, respectively). MILOP also provides at least 12 hours of warning before observed transfer in a greater proportion of severe episodes (53.7\% vs. 50.5\% and 46.9\%, respectively)---all while using 13.5\% and 24.6\% less combined Watcher and CCOT patient-time than the two baselines, respectively.

\subsection{Related Literature}\label{sec: lit review}
{\color{black}
This research is connected to several streams of literature related to RMABs, learning through optimization, and healthcare decisions under limited resources. 

\paragraph{RMABs and constrained decision processes.}
RMABs model systems in which individual processes continue to evolve even when they are not selected for intervention \citep{whittle_restless_1988}. Because solving the joint decision problem is computationally intractable, much of the literature develops index policies and linear programming relaxations \citep{papadimitriou_complexity_1999,weber_index_1990}. Extensions to multi-action and heterogeneous bandits broaden the framework beyond binary activation \citep{hodge_asymptotic_2015,verloop_asymptotically_2016}. Some of this work considers multiple levels of intervention under a single overall resource constraint \citep{hodge_asymptotic_2015}, whereas our setting provides a separate budget for each escalation action. Specifically, Watcher surveillance, CCOT assessment, and proactive transfer recommendations draw on different resources, requiring the policy to determine both which patients receive an intervention and which intervention each patient receives. More generally, \citet{goldsztajn_asymptotically_2024} use occupancy measures and fluid limits to construct asymptotically optimal policies for weakly coupled Markov decision processes with multiple actions and constraints, under stated conditions. We build on this planning approach to develop a learned policy for a finite and changing ward population. In our setting, the occupancy plan specifies allocations under average resource limits, but these allocations need not be jointly feasible for the patients present at a particular decision time. To address this challenge, MILOP learns patient--action priorities from the occupancy plan and uses a constrained assignment problem to generate recommendations that satisfy the separate action budgets. As such, the occupancy plan informs the long-run allocation of resources, while the assignment problem ensures that recommendations are feasible for the current ward population.

\paragraph{Learning through optimization.}
Differentiable optimization allows a learning algorithm to account for the decisions made from its outputs \citep{amos2017optnet,agrawal2019differentiable}. Decision-focused learning uses this connection to align model fitting with downstream decisions \citep{donti2017task,wilder2019melding,mandi2024decision}, including sequential decisions \citep{amos2018differentiable,futoma2020popcorn}. Within the subset of literature focused on RMABs, \citet{wang2023scalable} differentiate a Whittle-index policy to learn transition dynamics. Learned indices can also support joint allocation: \citet{zamir_deep_2024} learn partial indices with actor--critic networks and use them in maximum-weight matching for multi-resource restless bandits. For more general combinatorial action constraints, \citet{xu_reinforcement_2025} embed a learned Q-network in mixed-integer optimization. Compared to these prior works, MILOP differs in that it learns action scores by imitating an occupancy plan through a differentiable allocator. This approach addresses the need to translate long-run planning allocations into feasible recommendations for a changing set of patients. Specifically, applying the occupancy plan's conditional action distributions independently across patients can exceed the available resources at a given decision time. MILOP accounts for this competition during training by evaluating the joint allocation produced by the learned scores against the occupancy targets. Consequently, the network learns priorities based on how its scores affect the allocation of resources across patients. The occupancy plan provides supervision for this learning stage, without requiring the network to learn transition dynamics or an action-value function. With learning focused on downstream allocation, the upstream risk predictor remains fixed, and a separate assignment problem converts the learned scores into discrete recommendations at execution. Our approach therefore connects occupancy-based planning with index learning and constrained allocation to support the multiple escalation decisions within CHOA's watcher process.

\paragraph{Healthcare decisions under limited resources.}
Constrained Markov decision models have been used to allocate resources in breast cancer diagnosis \citep{ayvaci2012budgetary} and sequential genetic testing \citep{adelman2024genetic}. In maternal and child health, RMAB policies allocate limited outreach across patients whose engagement evolves over time \citep{mate2022field,verma2023restless,wang2023scalable}. Ward deterioration presents a related allocation problem, but one involving several escalation options with different capacity units. Decisions depend on the recorded risk stratum, while the severity of a subsequent transfer provides an outcome against which to evaluate those decisions.

A related line of operations research combines prediction with resource allocation, for example by predicting donor-milk composition before optimizing donation pools \citep{chan2023milk} or learning delivery-time distributions for order fulfillment \citep{ye2025contextual}. In our setting, the risk prediction is an input to a policy that must choose among several actions. MILOP adds a learning stage in which priorities are fitted through an allocation problem. Establishing the value of that stage calls for comparisons with simpler risk rules, just as learned policies in other operations settings are assessed against established heuristics \citep{gijsbrechts2022inventory}.

Our work is also closely related to studies of proactive ICU transfer decisions. \citet{hu2018examination} examine early transfers guided by a physiologic risk score, combining empirical analysis with simulation to evaluate the potential benefits of earlier admission and the resulting effects on ICU congestion. \citet{grand-clement_robustness_2023} formulate proactive transfer decisions as an MDP and establish conditions under which optimal nominal and robust policies have a threshold structure. They also relate the single-patient problem to a multi-patient formulation with a shared transfer constraint. Similar to these studies, our framework uses patient-level risk information to support decisions before deterioration leads to an urgent transfer. However, our setting includes Watcher surveillance and CCOT assessment in addition to usual care and proactive transfer, with a separate budget for each escalation action. These additional actions can affect the patient's subsequent course while the patient remains on the ward. Accordingly, the decision extends beyond whether to transfer a patient to determining how limited surveillance, outreach, and transfer resources should be allocated across patients over time. A transfer threshold alone does not specify these assignments, motivating our multi-action RMAB formulation with heterogeneous budgets and the development of MILOP to coordinate the available resources.

\subsection{Contributions}\label{sec: contributions}
In light of the practical challenges faced by CHOA and the gaps in existing literature, our paper makes three technical and practical contributions.

First, we develop a multi-action RMAB formulation for jointly allocating heterogeneous ward-escalation resources. The model captures a setting in which a decision-maker periodically chooses among usual care, Watcher surveillance, CCOT assessment, and proactive ICU-transfer recommendations for multiple patients being monitored concurrently. Each escalation action is governed by its own resource constraint, reflecting distinct requirements such as staffing capacity and an allowance for proactive transfer recommendations. Each patient's state also evolves stochastically whenever they are not transferred to the ICU. Moreover, our reward function reflects the relative costs of each intervention compared to usual care and captures the effect of increasing patient risk for deterioration. Lastly, we parameterize this model using a combination of clinician input and historical data. 
Overall, our model can be generalized to one that captures the sequential allocation of several limited hospital resources across multiple stochastically evolving patients and the modeled consequences of those decisions for subsequent patient outcomes.

Our second contribution addresses the challenge of solving this multi-action RMAB model through \emph{MILOP}, the \emph{Multi-action Index Learning and Optimization Policy}. MILOP combines long-run planning with differentiable optimization to learn how to allocate actions jointly across the ward. We first solve an occupancy LP to plan resource use over time, accounting for how patients’ conditions may change under each action. The resulting targets may not be feasible for the patients currently on the ward. MILOP therefore uses a neural network to score each patient–action pair and a differentiable allocation layer to distribute resources across patients within the available budgets. During training, the network adjusts its scores to make the resulting allocations match the LP targets as closely as possible. We implement the layer with entropy-regularized optimal transport \citep{cuturi2013sinkhorn,cuturi2019differentiable,xie2020differentiable}. At execution, constrained assignment converts the learned indices into discrete recommendations that satisfy each action budget without requiring all available resources to be used. The resulting policy therefore incorporates long-run planning information while coordinating recommendations across patients competing for the same resources.

Finally, we operationalize and retrospectively evaluate the framework in the CHOA setting. We combine historical patient trajectories with clinician-informed model inputs, including explicit assumptions about the effects of escalation actions that are not directly identifiable from the observational data. We evaluate the resulting policy and compare it with two practically relevant benchmark allocation approaches. MILOP reaches $87.4\%$ of critical and emergent transfer episodes before ICU departure, compared with $79.5\%$ for the stronger benchmark, while using $13.5\%$ less combined Watcher and CCOT monitoring. MILOP provides a median lead time of $19.2$ hours, with paired median lead-time advantages of $1.0$ and $5.5$ hours relative to the two benchmarks. Its recommendations also exhibit clinically intuitive staged escalation: $99.3\%$ of MILOP's proactive ICU-transfer recommendations are preceded by Watcher surveillance or CCOT assessment.

Together, these contributions provide a framework for moving from patient-level deterioration risk to repeated, resource-feasible escalation decisions. The model distinguishes among clinically different responses rather than treating escalation as a single binary intervention, and the solution approach connects long-run system planning with decisions that can be implemented at each decision epoch. Beyond the CHOA application, the framework applies to settings in which several limited interventions must be allocated among individuals whose needs evolve over time.

The remainder of this paper is organized as follows. Section~\ref{sec:setting} describes the care setting at CHOA, the escalation process, and the data. Section~\ref{sec:choa_model} develops the RMAB model for joint patient monitoring and resource allocation, and Section~\ref{sec:choa_nip} presents MILOP. Section~\ref{sec:parameterization} describes the model parameterization, and Section~\ref{sec:experiments} evaluates MILOP's recommendations using historical CHOA data, focusing on coverage, recommendation timing, and resource use. Section~\ref{sec:proposed-managerial} draws managerial insights from these results and followed by Section~\ref{sec: implementation} describing implementation plan. Finally, Section~\ref{sec:proposed-conclusion} summarizes the contributions and proposed future research direction.

\section{Children's Healthcare of Atlanta (CHOA) Setting}
\label{sec:setting}
At CHOA, care teams must choose among several escalation responses using risk observations that arrive intermittently. We first describe that care process and the available data (i.e., electronic health records), which together determine the information and resource constraints represented in our modeling framework of Section \ref{sec:choa_model}. Throughout, we distinguish quantities estimated from the data with capacities and intervention assumptions informed by clinical discussions.

\subsection{Practice Partnership and Translation}
\label{sec:translation}
This work sits within Pillar 1 of the CHOA Pediatric Technology Center, a partnership that brings clinicians together with Georgia Tech researchers and engineers. Pillar 1 encompasses several projects, among which the wider clinical deterioration effort includes a project named the Algorithm for Counteracting the Clinical Onset of Rapid Deterioration in Pediatric Patients (ACCORD). The ACCORD project is simultaneously developing models, applications, and user testing and its technical work combines secure electronic health record data, model development, and computational infrastructure for integrating outputs into hospital workflows \citep{gatech2026ipat}. 
Key stakeholders are intentionally incorporated in various design and implementation steps along the way. For example, frontline nurses have participated in several iterative design stages, and integration into CHOA's electronic health record is planned in the coming months with support from CHOA's informatics core \citep{gatech2026accord}. Deployment planning is also underway within the program. This institutional setting provides a route for evaluating MILOP's recommendations with the people and systems that would act on them. 

\subsection{Escalation Practice, Resources, and Coordination}
\label{sec:practice}
Within the watcher process being designed at CHOA, four responses are available to a ward team. (i) \emph{Usual care}: this comprises routine nursing and physician assessment, including clinicians' existing escalation decisions. (ii) \emph{Watcher}: when a patient warrants closer surveillance and evaluation, the team may designate the child a \emph{Watcher} \citep{evans2021implementing}. (iii) \emph{CCOT}: patients at elevated risk may also receive assessment by the \emph{critical care outreach team} (CCOT), whose involvement clinicians describe as lasting at least 12 hours. (iv) \emph{ICU Transfer}: a proactive \emph{ICU transfer} seeks admission before urgent rescue becomes necessary. These responses draw on distinct clinical resources and escalation mechanisms that already exist within the hospital, but historically they have not been coordinated through a single ward-level decision process. Rather, each has operated through its own clinical workflow and resource constraints. The watcher process being designed at CHOA brings these previously separate responses into a common escalation framework in which the team can jointly consider which patient should receive which level of response and when.

\section{Ward Escalation Model}
\label{sec:choa_model}


In this section, we describe our multi-action RMAB model for ward escalation. In our modeling framework, we assume a finite pool of $M$ patients that are being concurrently monitored. We model each patient $m=1,...,M$ as a Markov Decision Process $(\mathcal{T}, \mathcal{S}, \mathcal{A}, P, r)$.Figure~\ref{fig:ward-escalation} provides an overview of
the ward states, available responses, and possible next
states. We define the model components below.

\begin{figure}[t]
\centering
\FIGURE
{\includegraphics[width=\linewidth]
  {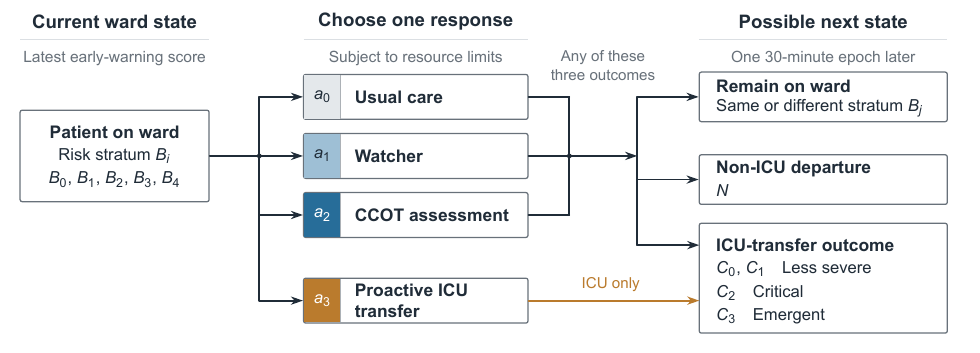}}
{Ward escalation decisions and possible next states.
\label{fig:ward-escalation}}
{At each 30-minute decision epoch, a patient in ward in $B_i$ is assigned one response. Usual care ($a_0$), Watcher surveillance ($a_1$), and CCOT assessment ($a_2$) each permit continued ward care in stratum $B_j$, non-ICU departure ($N$), or an ICU-transfer
outcome ($C_0,\ldots,C_3$). Proactive ICU transfer ($a_3$) leads only to an ICU-transfer outcome.}
\end{figure}

\paragraph{Decision epochs.} Let $\mathcal{T} = \{0,1, 2, ... \}$ be the set of decision epochs, which we model to be infinite and regularly spaced with decisions occurring every 30 minutes, or 48 times per day. That is, epochs $0,..., 47$ correspond to day 1, epochs $48, ..., 95$ correspond to day 2, and so on. In general, for any non-negative integer $\delta \in \mathbb{Z}_{\geq0}$, the epochs $48 \delta, ..., 48 (\delta + 1) - 1$ correspond to day $\delta + 1$.

\paragraph{States.} Let $S_{mt} \in \mathcal{S}$ denote the random state of patient $m$ in decision epoch $t$, where $\mathcal{S} = \mathcal{B} \cup \mathcal{E}$ is the set of states. The set $\mathcal B=\{B_0,\ldots,B_4\}$ represents the set of ward states in which each state represents a range of early warning scores. Ward states are ordered such that the predicted risk of deterioration for patients in state $B_i$ are greater than those in $B_{i-1}$, $i=1,..,4$. The set $\mathcal E=\{N,C_0,C_1,C_2,C_3\}$ represents
ward-exit outcomes. State $N$ denotes departure from the ward without an ICU transfer. States $C_0,\ldots,C_3$
distinguish ICU transfers by intervention type and timing. State $C_3$ denotes an \emph{emergent} transfer involving intubation or vasoactive support before transfer or within the first hour after transfer. State $C_2$ denotes a \emph{critical} transfer that does not meet the emergent definition but involves noninvasive positive-pressure ventilation (NIPPV) within 12 hours of transfer, or intubation or vasoactive support 1--12 hours after transfer. Among the remaining transfers, $C_1$ denotes those with the first intervention occurring more than 12 hours and no more than 24 hours after transfer, and $C_0$ denotes all remaining ICU transfers. Our clinical collaborators at CHOA provided the emergent and critical transfer definitions.

\paragraph{Actions.} In each decision epoch $t$, our model recommends an action $A_{mt} \in \mathcal A=\{a_0,a_1,a_2,a_3\}$ for each patient $m=1,...,M$, wherein $a_0$ represents usual care, $a_1$ represents labeling a patient as a Watcher, $a_2$ represents sending the CCOT, and $a_3$ represents performing a proactive ICU transfer. Because usual care already includes existing clinical escalation, the Watcher and CCOT actions represent additional interventions relative to that baseline. When a patient is in an outcome state $s \in \mathcal{E}$, no further action is possible. Thus, for notational convenience, we denote this constraint by choosing $a_0$. Lastly, each actions $a_i, i \ne 0$ has a separate capacity constraint that must be satisfied in each day. That is, $\sum_{m=1}^M \Ind{A_{m t} = a_i} \leq b_i$ for $i = 1,2,3$ and for any epoch $t \in \mathcal{T}$, where $\Ind{\cdot}$ is the indicator function.

\paragraph{Transition probabilities.} For each patient $m=1,...,M$, we model transition probabilities as $P^a(s,s') = \Pr(S_{m,t+1} = s'|S_{mt} = s,A_{mt} = a)$ denotes the probability of transitioning from state $s$ to state $s'$ over one decision epoch under action $a$. Under usual care, Watcher, and CCOT actions, a patient may move between risk strata or leave the ward through a non-ICU exit or an ICU transfer of varying severity. That is, $P^{a}(s, s') > 0$ for any $a\in \{a_0, a_1, a_2\}$ and $s, s' \in \mathcal{S}$. These transition probabilities differ in their likelihood of reaching each outcome state $s' \in \mathcal{E}$. In particular, both Watcher $(a_1)$ and CCOT $(a_2)$ actions shift probability from the critical transfer $(C_2)$ and emergent transfer $(C_3)$ outcome states compared to usual care $(a_0)$, with this shift being more pronounced when $a_2$ is chosen compared to $a_1$. When a proactive ICU transfer ($a_3$) is chosen, the patient may only transition to one of the ICU transfer outcome states $C_0-C_3$. That is, $P^{a_3}(s, s') = 0$ for $s' \notin \{C_0, ..., C_3\}$. Finally, once any patient reaches an exit state $s \in \mathcal{E}$, our modeling framework assumes that they are replaced with a new patient in the next transition. Thus, the next state occupied is a ward state. That is, for any $s \in \mathcal{E}$, $P^{a_0}(s, s') = 0$ for any $s' \notin \mathcal{B}$. 

\paragraph{Rewards.} Let $r(s_{mt},A_{mt})$ denote the reward incurred by the model when patient $m$ is in state $S_{mt}$ and action $A_{mt}$ is taken in decision epoch $t$. Based on discussions with clinical experts at CHOA, we designed the reward function to penalize critical and emergent transfers to the ICU while accounting for the burden of remaining at high acuity and using escalation resources. That is,

for any outcome states $s \in \mathcal{E}$, we assume that $r(N,a_0)=r(C_0,a_0)=r(C_1,a_0)=0$ and $r(C_3, a_0) \leq  r(C_2, a_0) \leq 0$. Before an exit state is reached, patients in a ward state $s = B_i$, $i=0,...,4$, receiving action $a_j, j=0,1,2,3$ incur a linearly additive reward of $r(s_i, a_j) = -(u_i + c_j)$. To reflect the relative costs associated with each escalating ward state, we assume that the disutility $u_i$ associated with being in ward state $B_i$ is non-decreasing in $i$, i.e., $u_{i+1} \geq u_i$ for $i=0,1,2$. Likewise, we capture the increasing resource strain associated with each action by assuming that the disutility associated with each action is non-decreasing in $j$, i.e., $c_{j+1} \geq c_j$ for $j=1,2,3$. 

\paragraph{Objective.} The model's objective is to maximize long-run average rewards subject to resource constraints by identifying an optimal policy that specifies which action to recommend for each patient based on the states of all patients. Mathematically, this optimization problem is given by 
\begin{subequations}\label{eq:choa_full_problem} 
\begin{align}
        \sup_{\pi \in \Pi} g(\pi) &:=  \liminf_{T \to \infty} \frac{1}{MT} \E_{\pi} \left[ \sum_{t=0}^{T-1}\sum_{m=1}^M r(S_{mt}, A_{mt}) \right] \\
    \text{s.t.} \quad & \sum_{m=1}^M \Ind{A_{mt} = a_i} \leq b_i \text{ for all $i=1,2,3$, $t \in \mathcal{T}$}, \label{eq:fullmodel_constraint1} 
\end{align} \end{subequations}
where $\pi: \mathcal{S}^M \to \mathcal{A}^M$ denotes a deterministic policy that specifies a recommended action for each patient given the states of all patients, $\Pi$ is the set of all Markovian deterministic policies, and the expected value $\E_{\pi}[\cdot]$ takes the expectation over all random state trajectories under policy $\pi$. We anticipate that the optimization problem \eqref{eq:choa_full_problem} is computationally intractable since the classic RMAB problem with only a single resource constraint and two actions are known to be PSPACE-hard in general \citep{papadimitriou_complexity_1999}. Compared to the classic RMAB setting, \eqref{eq:choa_full_problem} is also complicated by the fact that constraints are pathwise (i.e., must be satisfied in each sample path). Accordingly, we develop MILOP to obtain a high-quality solution to \eqref{eq:choa_full_problem} within a computationally feasible timeframe for CHOA.

\choahead{section}{MILOP: Multi-action Index Learning and Optimization Policy}
{\color{black}
\label{sec:choa_nip}\label{sec: Methods}
We now present MILOP, our algorithmic approach for generating feasible and high-quality solutions to \eqref{eq:choa_full_problem}. At a high level, MILOP is a decision-focused learning framework that leverages neural networks to predict multi-action indices to guide decisions in the watcher process while respecting capacity constraints (see Figure \ref{fig:pipeline}). Critically, a knapsack formulation is used to transform predicted index scores into recommended actions. However, this knapsack formulation is not amenable to training the neural network. As such, a key innovation in MILOP is the use of a differentiable optimal transport problem during the training process that is used to approximate the knapsack problem. MILOP then guides the neural network through a decision-focused loss function that minimizes the discrepancy between the neural network's predicted actions and the optimal occupancy measure derived from an LP relaxation of \eqref{eq:choa_full_problem}. We next describe this approach in more detail. 

}

\begin{figure*}[t]
\centering
\FIGURE
{\includegraphics[width=\linewidth]{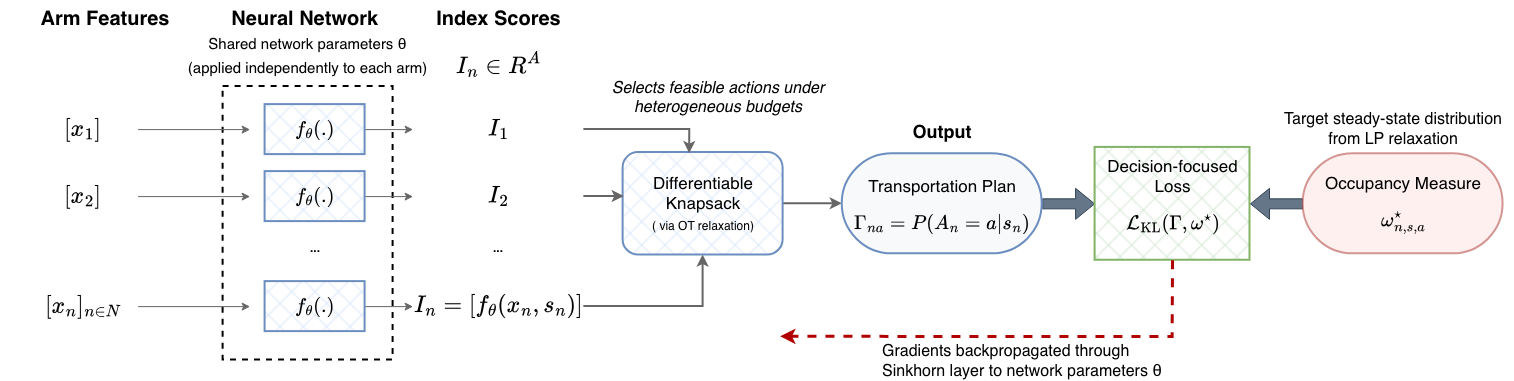}}
{MILOP architecture.\label{fig:pipeline}}
{The same network produces action scores for every patient.
Training compares the row-normalized joint allocation with
conditional LP targets. The dashed arrow shows backpropagation
through the transport layer.}
\end{figure*}

\subsection{Neural Network for Index Prediction}

In MILOP, we leverage a neural network to jointly predict a multi-action index score for each patient. Specifically, let $f_{\theta}$ denote this neural network with parameters $\theta$ and let $\bm{x}_m$ denote a vector of patient-specific features. Then, the multi-action index is given by a vector $I_{m} = [I_{ma}]_{a \in \Actions} = f_\theta (\bm{x}_m)$, wherein each index score $I_{ma}$ represents the priority of recommending action $a$ for patient $m$. The feature vector $\bm{x}_m$ encodes the patient's current ward state $S_{mt}$ together with the latest early warning score, elapsed ward time, and recent score history, using only information recorded at or before epoch $t$. The state therefore enters the network through $\bm{x}_m$ rather than as a separate argument. E-Companion~\ref{ec:network_config} specifies the nine inputs, the network architecture, and the training settings. We remark that, while index predictions are made on each patient through individual application of the neural network $f_\theta$, we couple all patients together in the training process by updating parameters $\theta$ based on our decision-focused training loss that accounts for the joint decisions across all patients.  

Given multi-action index vectors $I_m$ for all $m=1,...,M$, MILOP next aims to determine action recommendations for each patient that are feasible in \eqref{eq:choa_full_problem}. In the classic RMAB, a ``top-$k$" approach can be used with index scores to identify which arms should be pulled. However, our multi-action setting necessitates a multi-action index vector $I_m$ for each patient, which makes the classic approach difficult to apply.

Because the index scores $I_{ma}$ correspond to the importance of recommending action $a$ for patient $m$, we model the transformation of index scores to feasible actions as a knapsack problem \begin{subequations} \label{eq: knapsack problem}
    \begin{align}
 \max_{z}\quad &\sum_{m=1}^{M}\sum_{a\in\mathcal A} I_{ma}z_{ma},\\
 \text{s.t.}\quad
 &\sum_{a\in\mathcal A}z_{ma}=1 \text{ for $m=1,...,M$},\label{eq:budget_const}\\
 &\sum_{m=1}^{M}z_{ma_i}\leq b_i \text{ for $i=1,2,3$}\\
 &z_{ma}\in\{0,1\} \text{ for all $m=1,...,M$, $a \in \Actions$}.
\end{align}
\end{subequations}
wherein the decision variable $z_{ma}$ equals $1$ if action $a$ is recommended for patient $m$ and is equal to $0$ otherwise. Clearly, an optimal solution $z^*$ to \eqref{eq: knapsack problem} must also be feasible in \eqref{eq:choa_full_problem}. We next describe how we train our neural network $f_\theta$ to output index scores that lead to high-quality ward escalation decisions.

\subsection{Training the Neural Network}
Our training process involves using the decision policy generated by the neural network $f_\theta$, evaluating the performance of the policy, and then using backpropagation to end-to-end update the parameters $\theta$ to improve the intervention performance. However, we face two key challenges in this training process. The first challenge is that the knapsack problem in \eqref{eq: knapsack problem} is not differentiable because of the binary constraints on $z$. The second challenge is that we need to derive a training loss function that leads to high-quality decisions. We next describe our algorithmic design decisions to overcome these challenges. 

\subsubsection{An Optimal Transport Problem. }Since \eqref{eq: knapsack problem} is non-differentiable, we instead use a differentiable optimal transport problem to approximate \eqref{eq: knapsack problem}. That is, this optimal transport problem considers the mass transport between a source distribution representing each patient and a target distribution representing the capacity constraints. Specifically, let $\Gamma \in \mathbb{R}^{(M+1) \times |\Actions|}$ denote the transport plan which is equivalent to a matrix wherein each column $a \in \Actions$ corresponds to an action, rows $m=1,...,M$ correspond to patients, and row $M+1$ corresponds to unused capacity. The optimal transport plan seeks to identify a transport plan that solves
\begin{subequations}\label{eq: OT problem new}
    \begin{align}
        \min_{\Gamma \geq 0} \quad & \sum_{m=1}^{M+1} \sum_{a \in \Actions} -I_{ma} \Gamma_{ma} + \epsilon \Gamma_{ma} \left( \log \Gamma_{ma} - 1 \right) \label{eq: OT objective} \\
        \text{s.t.} \quad & \sum_{a \in \Actions} \Gamma_{ma} = 1 \text{ for $m=1,...,M$} \label{eq: OT constraint 1}\\
        & \sum_{a \in \Actions} \Gamma_{M+1, a} = b_1 + b_2 + b_3 \label{eq: OT constraint 2}\\
        & \sum_{m=1}^{M+1} \Gamma_{m, a_0} = M \label{eq: OT constraint 3}\\
        & \sum_{m=1}^{M+1} \Gamma_{m, a_i} = b_i \text{ for $i=1,2,3$}.  \label{eq: OT constraint 4}
    \end{align}
\end{subequations}
In \eqref{eq: OT problem new}, the first term in the objective function \eqref{eq: OT objective} maximizes the value associated with $\Gamma$ (i.e, by maximizing allocation across index scores), the second term in the objective is the negative entropy term, and the parameter $\epsilon > 0$ controls the smoothness of the resulting transport plan. That is, as $\epsilon$ approaches $0$, the resulting transport plan appears more ``discrete". Constraints \eqref{eq: OT constraint 1}-\eqref{eq: OT constraint 4} ensure that the resulting policy generated by the optimal transport plan $\Gamma^*$ does not violate the resource constraints for actions $a_i, i=1,2,3$. The constraints assign one unit of mass to each patient, which can be divided among actions. The additional row absorbs unused capacity, ensuring that the total patient allocation to each budgeted action does not exceed its resource limit. To facilitate the computational efficiency of our training process, we solve \eqref{eq: OT problem new} using the Sinkhorn scaling algorithm \citep{cuturi2013sinkhorn}. This algorithm alternately rescales rows and columns to match the totals specified by the constraints. In our implementation of this algorithm, we perform all calculations in the log domain for numerical stability. 
Finally, we can convert any optimal transport plan $\Gamma^*$ to a policy that picks action $a$ for patient $m$ with probability \begin{equation}
    \widehat{\Gamma}_{ma} := \frac{ \Gamma_{ma} }{ \sum_{a' \in \Actions} \Gamma_{ma'}} = \Pr(A_{mt} =a| S_{mt}). \label{eq: OT randomized policy}
\end{equation}



\subsubsection{Decision-focused Training Loss. }During training, we need a way to guide the neural network towards predicting index scores $(I_{m})_{m=1}^M$ that lead to high-quality ward escalation decisions. In the ideal case, we would be able to use the optimal objective function value to \eqref{eq:choa_full_problem} compare against the value of the policy generated by MILOP. However, solving \eqref{eq:choa_full_problem} is challenging for many reasons, including the scale of the optimization problem and the need to impose constraints \eqref{eq:fullmodel_constraint1} at each decision epoch. To this end, we next describe our approach to generating a training target based on an LP relaxation of \eqref{eq:choa_full_problem} in which the capacity constraints must only be satisfied in expectation. This LP relaxation can then serve as a guide for MILOP while training the neural network towards generating high-quality policies. 

Relaxing the constraints in \eqref{eq:choa_full_problem} so that the constraints are only satisfied in expectation yields the following optimization problem:
\begin{equation} 
    \sup_{\pi \in \Pi} \left\{g(\pi): \lim_{T\to\infty} \frac{1}{T} \sum_{t=0}^{T-1} \sum_{m=1}^M \Ind{A_{mt} = a_i} \leq b_i \text{ for $i=1,2,3$} \right\}. \label{eq:choa_full_problem_relaxation}
\end{equation}
The optimization problem \eqref{eq:choa_full_problem_relaxation} can be solved using an occupancy-based formulation, similar to what has been developed in prior work for weakly coupled MDPs and RMABs \citep{altman2021constrained, verloop_asymptotically_2016,goldsztajn_asymptotically_2024}: 
\begin{subequations}\label{eq:choa_lp}
\begin{align}
\omega^* =\argmax_{\omega\geq0}\quad&\sum_{s \in \States} \sum_{a \in \Actions} \omega(s,a)r(s,a),\\
\text{s.t.}\quad&\sum_{a \in \Actions} \omega(s,a)=\sum_{s' \in \States} \sum_{a' \in \Actions} \omega(s',a')P^{a'}(s',s) \text{ for all $ s \in \States$} \\
&\sum_{s \in \States} \sum_{a \in \Actions}\omega(s,a)=1,\\
&\sum_{i=0}^{4}\omega(B_i,a_j)\leq b_j/M,\quad j=1,2,3\\
&\omega(s,a)=0,\quad s\in\mathcal{E},\ a\ne a_0.
\end{align}
\end{subequations}

In \eqref{eq:choa_full_problem_relaxation}, the decision variables $\omega(s,a)$ are known as occupancy measures. If we take the policy $\pi^\omega$ to be a randomized policy such that for every $m=1,..,M$, we pick action $a$ in state $s$ with probability \begin{equation}
    \pi^\omega(a|s) := \frac{ \omega(s,a)}{\sum_{a' \in \Actions} \omega(s,a')}, \label{eq: policy pi omega}
\end{equation} then we can interpret the occupancy measure as the long-run proportion of time that any patient spends in state $s$ and has action $a$ recommended to them \citep{altman2021constrained}. That is, $\omega(s,a) = \lim_{T \to \infty} \frac{1}{T} \E_{\pi^\omega}[ \sum_{t=0}^{T-1} \sum_{m=1}^M \Ind{S_{mt} = s, A_{mt} = a}]$. 
Let
\[
L^\star
=
\sum_{s\in\States}\sum_{a\in\Actions}
\omega^\star(s,a)r(s,a)
\]
denote the optimal value of the occupancy-measure LP in \eqref{eq:choa_lp}. For every policy $\pi\in\Pi$ that is feasible for the original problem \eqref{eq:choa_full_problem}, we have $g(\pi)\leq L^\star$, so the optimal value of \eqref{eq:choa_full_problem} is also bounded above by $L^\star$. Thus, \eqref{eq:choa_lp} is a relaxation of the original problem and provides an upper bound on the performance achievable by any feasible policy. Importantly, this bound applies to the optimal LP value rather than to the value $g(\pi^{\omega^\star})$ realized by the randomized policy induced by $\omega^\star$, since $\pi^{\omega^\star}$ need not satisfy the pathwise resource constraints in \eqref{eq:fullmodel_constraint1}. We therefore use the optimal LP occupancy measure $\omega^\star$ as the training target for MILOP rather than directly implementing $\pi^{\omega^\star}$. Specifically, MILOP's training loss compares the policy induced by the optimal transport plan $\Gamma^\star$ with the policy induced by $\omega^\star$ using the Kullback--Leibler (KL) divergence.: 
\begin{equation}
    \mathcal{L}_{\text{KL}}(\Gamma^*, \omega^*) = \sum_{m=1}^M \sum_{a \in \Actions} \pi^{\omega^*}(a|S_{mt}) \log \frac{\pi^{\omega^*_{m}}(a|S_{mt})}{\widehat{\Gamma}_{ma}}, \label{eq: training loss}
\end{equation}
where $\pi^{\omega^*}$ is defined in \eqref{eq: policy pi omega} and $\widehat{\Gamma}_{ma}$ is defined in \eqref{eq: OT randomized policy}. Minimizing this loss encourages MILOP to produce allocations that more closely match the LP targets. During training, the target policy $\pi^{\omega^*}$ remains fixed, while the allocation probabilities $\widehat{\Gamma}$ depend on the network parameters $\theta$ through the predicted index scores and the transport problem. Backpropagation computes how each network parameter affects the loss by differentiating through the row normalization and the Sinkhorn iterations. The optimizer then uses these gradients to update $\theta$ with the aim of reducing the discrepancy between the allocation probabilities and the LP targets. Since the transport problem allocates shared resources jointly across patients, changing one patient's index scores can also change the allocations to other patients. Through repeated updates, the network therefore learns patient--action priorities based on their consequences for the joint allocation, accounting for competition for limited resources.


\subsubsection{Training Algorithm.} We now describe our full training algorithm, which is summarized in Algorithm \ref{alg:training new}. As input to the algorithm, we assume that the MDP models are fully parameterized and that we have time series data that are divided (temporally) into training and validation datasets. The first step in our algorithm is to solve \eqref{eq:choa_lp} in order to obtain the optimal occupancy measures $\omega^*$ which are used to obtain the policy $\pi^{\omega^*}$ that guides MILOP. The resulting target distributions $\pi^{\omega^*}$ remain fixed throughout training.  We also initialize the neural network $f_\theta$. Then, within each training epoch, we sample decision times
from the training period and include all patients eligible
for recommendations at each selected time. For each patient,
we determine the state $S_{mt}$ from the early warning score and construct the feature vector $\bm{x}_m$, which encodes that state alongside the remaining patient-specific measures. Then, we apply the neural network and obtain index scores $I_m = f_\theta (\bm{x}_m)$ for each patient $m=1,...,M$. Using these index scores, we obtain the optimal transport plan $\Gamma^*$ by solving \eqref{eq: OT problem new} and derive the policy $\hat{\Gamma}$ according to \eqref{eq: OT randomized policy}. We then compute the training loss \eqref{eq: training loss} and update $\theta$ by backpropagation. At scheduled checks, we apply the same allocation and loss
calculations to the validation data without updating
$\theta$. We compute the validation loss against the fixed
LP targets and retain the parameters $\theta^*$ that achieve
the lowest validation loss.

\begin{algorithm}
\color{black}
\caption{MILOP Training Algorithm}\label{alg:training new}
\begin{algorithmic}[1]
\State \textbf{Input:} Training and validation data, model inputs, capacities
\State Solve \eqref{eq:choa_lp}, obtain $\omega^*$ and initialize $f_\theta$
\For{each training epoch}
 \For{each training data point}
  \State Determine states $S_{mt}$ and construct patient features $\bm{x}_m$ for all $m=1,...,M$
  \State Compute scores $I_{m}=f_\theta(\bm{x}_m)$ and optimal transport plan $\Gamma^*$ by solving \eqref{eq: OT problem new}
  \State Form both $\pi^{\omega^*}$ and $\widehat{\Gamma}$ and evaluate \eqref{eq: training loss}
  \State Update $\theta$ by backpropagation
 \EndFor
 \State At scheduled checks, evaluate validation loss and retain the best parameters $\theta^*$
\EndFor
\State Report marginal errors and agreement with target-based assignment separately
\end{algorithmic}
\end{algorithm}

\subsection{Deploying MILOP}
We now describe how MILOP should be applied in practice. Specifically, given a trained neural network $f_{\theta^*}$ and a set of $M$ patients who are being concurrently monitored, the main difference from Algorithm \ref{alg:training new} is that the solution to the knapsack problem \eqref{eq: knapsack problem} is used to determine the recommended actions given computed index scores $[I_m]_{m=1}^M = [f_{\theta^*}(\bm{x}_m)]_{m=1}^M$ instead of the optimal transport plan. That is, patient $m$ is recommended action $z_{ma}$ if $z_{ma} = 1$. The watcher team at CHOA would then consider the recommendations, cross-check against real resource availability, and then make final decisions based on their own judgment.

\subsection{Day-wise Constraints and Practical Implementation} \label{sec:day-wise constraints}
We conclude this section with remarks on day-wise constraints. That is, in practice, the resource constraints in \eqref{eq:fullmodel_constraint1} may be better expressed on a daily basis even if it is desirable to keep decision epochs at every 30 minutes. This optimization problem is thus given by 
\begin{equation}
\sup_{\pi \in \Pi} \left\{ g(\pi): \sum_{t=0}^{47} \sum_{m=1}^M \Ind{A_{m, 48 \delta + t} = a_i} \leq b_i \text{ for all $i=1,2,3$, $\delta \in \mathbb{Z}_{\geq 0}$} \right\}. \label{eq: day-wise constraint}     
\end{equation}
It turns out that MILOP can accommodate \eqref{eq: day-wise constraint} with some minor changes. First, we note that relaxation of \eqref{eq: day-wise constraint} is given by \begin{align}
    &\sup_{\pi \in \Pi} \left\{ g(\pi): \lim_{\Delta \to \infty} \frac{1}{\Delta} \sum_{\delta=0}^{\Delta-1} \sum_{t=0}^{47} \sum_{m=1}^M \Ind{A_{m, 48 \delta + t} = a_i} \leq b_i \text{ for all $i=1,2,3$} \right\} \\
    =& \sup_{\pi \in \Pi} \left\{ g(\pi): \lim_{T \to \infty} \frac{1}{T} \sum_{t=0}^{T-1} \sum_{m=1}^M \Ind{A_{m, t} = a_i} \leq b_i/48 \text{ for all $i=1,2,3$} \right\}, 
\end{align}
which is equivalent to \eqref{eq:choa_full_problem_relaxation} with the right-hand side of the constraint scaled by $48$. Thus the training the process remains largely unchanged, with the key difference being in ensuring that the right-hand side of the constraint is scaled to a per-epoch level when solving the occupancy measure formulation \eqref{eq:choa_lp} and the optimal transport problem \eqref{eq: OT problem new}. 

Additionally, the deployment process must be slightly changed when considering day-wise constraints. Specifically, at any epoch $t = 48 \delta + k$, our remaining resources are given by $\tilde{b}_i(t) = b_i - \sum_{k'=0}^k \Ind{A_{m, 48 \delta + k'} = a_i}$ to ensure that we satisfy day-wise constraints.
We remark that our numerical study in Section \ref{sec:choa_case_study} reflects day-wise constraints which incorporate these modifications.

\section{Model Parameterization}
\label{sec:parameterization}

Having presented MILOP, we now describe how we parameterize the underlying ward escalation model for CHOA. Specifically, we combine estimates from historical ward records with clinical input to specify the transition probabilities $P^a(s,s')$, rewards $r(s,a)$, and resource limits. These inputs define the reference planning scenario. Table~\ref{tab:instantiation} summarizes the parameters and their sources, and E-Companion~\ref{app:choa_details} provides the detailed specification.
\begin{table}[h!]
\color{black}
\TABLE
{Sources of CHOA model inputs.\label{tab:instantiation}}
{\begin{tabular}{
>{\raggedright\arraybackslash}p{0.3\textwidth}
>{\raggedright\arraybackslash}p{0.70\textwidth}}
\hline\up
Model inputs & Basis \down\\
\hline\up
Transition inputs, $P^{a_0}$, $\nu$, and $H_i(c)$
& Estimated from training-period ward records \\

Intervention effects, $\rho_{\ell k}$ and $h$
& Assumptions informed by clinical discussions \\

Rewards, $r(s,a)$
& Clinician-reported mortality differences and assumed
ward and intervention costs \\

Resource limits, $b_1$, $b_2$, and $f_3$
& Clinical input and calibration to historical transfer
activity \down\\
\hline
\end{tabular}}
{}
\end{table}

\subsection{CHOA Cohort and Outcome Definitions}
\label{sec:choa_cohort}

Our historical data from CHOA cover January 1, 2017 through September 30, 2024. The data include ward admission and departure times, ICU-transfer times, the timing of respiratory and vasoactive support, and early warning scores. The early warning score was developed in a separate arm of the ACCORD project \citep{anonymous_pediatric_icu_transfer}. It uses information from the electronic health record to estimate the patient's risk of a critical or emergent ICU transfer within the next 12 hours.

We match each risk score observation to a ward stay using the encounter identifier and the recorded admission and departure times, excluding observations that cannot be matched. Of the initial 372,178 ward stays, we exclude 8,739 without a recorded risk score because the policy requires a score to recommend an action. We also exclude 106 stays ending in death on the ward without an ICU transfer, as ward death is not an outcome represented in the model. These deaths account for less than 0.03\% of the original sample. Table~\ref{tab:cohort} summarizes the final analysis sample. Of these ward stays, 354,767 (97.64\%) end without an ICU transfer, while 8,566 (2.36\%) end in an ICU transfer.

We classify ICU transfers into four mutually exclusive categories based on the type and timing of intervention. Our clinical collaborators at CHOA provided the definitions of emergent and critical transfers. Among ICU transfers, 5,667 (66.16\%) fall in category $C_0$, 279 (3.26\%) in $C_1$, 2,111 (24.64\%) in the critical category $C_2$, and 509 (5.94\%) in the emergent category
$C_3$. These outcome categories are defined in Section~\ref{sec:choa_model}.


Analyses provided by our clinical collaborators also show that in-hospital mortality following transfer differs across these categories. To distinguish the most severe transfer outcomes, we report critical and emergent transfers separately and jointly when evaluating the policies. Outcome classification uses intervention information recorded before and after transfer, while recommendations depend only on information available during the ward stay. 

\begin{table}[t]
\color{black}
\TABLE
{CHOA analysis data sample\label{tab:cohort}}
{\begin{tabular}{lr}
\hline\up
Measure & Count \down\\
\hline\up
Ward stays & 363{,}333\\
Hospital encounters & 304{,}098\\
Patients & 188{,}227\\
Recorded risk score observations & 13{,}516{,}821\\
Ward observation time (patient-days, rounded) & 919{,}012 \down\\
\hline
\end{tabular}}
{}
\end{table}



\subsection{Patient states and usual-care transitions}

We use the training-period data from CHOA to estimate the cutoffs defining the risk groups $B_0,\ldots,B_4$, the usual-care transition probabilities $P^{a_0}$, and the admission distribution $\nu$. Specifically, we determine the four cutoffs from empirical quantiles of the recorded early warning scores and hold them fixed throughout the subsequent analysis.

Let $p_i(c)$ denote the usual-care probability of departing from ward state $B_i$ to outcome $c\in\mathcal E$ during one decision epoch. We estimate these probabilities using observed departures and eligible ward time. Each ICU transfer is classified by its observed outcome and assigned to the risk group corresponding to the patient's most recent early warning score before transfer. For each risk group $B_i$ and transfer category $c\in\{C_0,C_1,C_2,C_3\}$, we divide the number of corresponding transfers by the eligible ward time in that group, measured in 30-minute epochs. We estimate $p_i(N)$ from non-ICU departures in the same way.

Since early warning scores are recorded at irregular intervals, we use the most recently recorded score to assign a patient's risk group between observations, limiting the time credited to each observation to 24 hours. Let $Q_{ij}$ denote the probability of moving from $B_i$ to $B_j$ conditional on remaining on the ward. We estimate $Q_{ij}$ from the relative frequencies of transitions between risk groups. The intervening epochs contribute to the probability of remaining in the same group, ensuring that movement and departure probabilities correspond to the model's 30-minute decision interval. Combining these estimates gives
\begin{equation}
P^{a_0}(B_i,c)=p_i(c),\quad c\in\mathcal{E},
\qquad
P^{a_0}(B_i,B_j)=\left[1-\sum_{c\in\mathcal{E}}p_i(c)\right]Q_{ij}.
\label{eq:usual_kernel}
\end{equation}

Finally, we estimate the admission distribution
$\nu=(\nu_0,\ldots,\nu_4)$, where $\nu_i$ is the proportion of episodes whose first recorded early warning score falls in $B_i$. Additional details on the risk cutoffs, treatment of longer observation gaps, and adjustments for sparse departure counts are provided in E-Companion~\ref{app:choa_details}.

\subsection{Intervention effects}

The data do not contain historical assignment labels for Watcher surveillance ($a_1$) or CCOT assessment ($a_2$) because a Watcher process was not previously implemented. Accordingly, we specify their effects through assumptions informed by clinical discussions. For each ward state $B_i$, we multiply the usual-care probabilities of critical and emergent transfer by action-specific factors: $ P^{a_\ell}(B_i,C_k) = \rho_{\ell k}P^{a_0}(B_i,C_k), \text{where }\ell=1,2, \text{and } k=2,3.$
Here, $\rho_{\ell k}$ is the proportion of the usual-care probability of outcome $C_k$ that remains under action $a_\ell$. In the reference scenario, we set
$(\rho_{12},\rho_{13})=(0.75,0.89)$ for Watcher surveillance and
$(\rho_{22},\rho_{23})=(0.60,0.67)$ for CCOT assessment.

We reassign the reduction in the probabilities of $C_2$ and $C_3$ to the less severe categories $C_0$ and $C_1$, according to their usual-care proportions. This preserves the overall probability of ICU transfer. The probabilities of non-ICU departure and movement between ward states remain unchanged. Thus, additional monitoring is assumed to support earlier care that reduces transfer severity without necessarily avoiding ICU admission.

For proactive transfer ($a_3$), let $H_i(c)$ denote the historical distribution of outcomes $c\in\{C_0,C_1,C_2,C_3\}$ among patients selected by clinicians for transfer from risk group $B_i$. We combine this distribution with an assumed handover interval of $h=2$ hours to specify $P^{a_3}(B_i,c)$. The handover interval determines the remaining probability of a critical or emergent outcome, while $H_i(c)$ determines how probability is divided within the severe and less severe categories. E-Companion~\ref{app:choa_details} provides the construction. These intervention effects are planning assumptions rather than estimates of causal effects.

\subsection{Rewards and intervention costs}

We calibrate exit penalties using clinician-reported in-hospital mortality rates of 1\%, 5\%, and 12\% following $C_1$, critical ($C_2$), and emergent transfers ($C_3$), respectively. Taking $C_1$ as the reference category yields $ r(C_2,a_0)=-4, r(C_3,a_0)=-11$, corresponding to mortality differences of four and eleven percentage points. We assign
$r(N,a_0)=r(C_0,a_0)=r(C_1,a_0)=0$
as a utility reference. Each exit reward is incurred once, during the exit-state epoch.

Additional costs penalize time spent in higher-risk ward states and the use of escalation resources. Let $u_i$ denote the cost of spending one decision epoch in ward state $B_i$, let $c_1$ and $c_2$ denote the Watcher and CCOT costs per patient-day, and let $c_3$ denote the cost of a proactive-transfer recommendation. The ward reward is
\begin{align}
    r(B_i,a)
    ={}&-u_i
    -\frac{c_1}{48}\mathbf{1}\{a=a_1\}
    -\frac{c_2}{48}\mathbf{1}\{a=a_2\}
    -c_3\mathbf{1}\{a=a_3\}.
\end{align}
Dividing $c_1$ and $c_2$ by 48 converts their daily costs to the model's 30-minute decision interval.

Together, these terms allow the planning model to balance severe outcomes against the burden of monitoring and intervention. The resulting reward is a planning objective and does not have a direct mortality or monetary interpretation. Table~\ref{tab:reward_params} reports the reference values and their units.

\subsection{Resource limits}
We translate clinical input and historical transfer activity
into resource limits for the planning model. Clinical input
guides our choice of $b_1=32$ concurrent Watcher assignments
and $b_2=25$ concurrent CCOT assignments. The Watcher limit
allows closer surveillance of approximately 10\% of the
reference ward census of $M=325$ patients. For proactive
transfer, we use the historical rate of 9,207 transfers over
2,830 days to set an average allowance of $f_3=3.25$
recommendations per day. Applying the modeling adjustments for day-wise resource constraints (see Section \ref{sec:day-wise constraints}), we divide this allowance across
48 daily decision epochs, giving
$\beta_3=f_3/48\approx0.0677$ recommendations per epoch.
These limits define resource availability for planning;
in Section~\ref{sec:eval_design}, we describe how the
retrospective evaluation applies rolling intake limits
and monitoring durations.

\section{Evaluation on Historical Data from CHOA}
\label{sec:experiments}

In this section, we evaluate MILOP using historical ward data from CHOA to assess how it allocates limited escalation resources and compare its performance against two baseline policies: \textit{top-$k$} and \textit{threshold-based-top-$k$}. Specifically, our numerical analysis evaluates which patients receive recommendations before a critical or emergent ICU transfer, how much time those recommendations allow for a response, and how much monitoring they require. 
We hold recorded patient trajectories and outcomes fixed throughout the evaluation. Thus, the analysis measures the allocation and timing of recommendations without estimating their effects on treatment decisions or subsequent deterioration.

\subsection{Baseline Policies}

We compare MILOP against two baseline policies called top-$k$ and threshold-based-top-$k$. These policies are inspired by index policies for classic RMAB problems and threshold policies that have been previously used in other ward escalation models (see Section \ref{sec: lit review}). The key difference between each baseline is in how they prioritize patients and assign the available escalation resources, as we next describe. 
Together, these baselines allow us to compare MILOP with risk-based allocation rules that differ in whether they require a minimum predicted risk before assigning an action.

\paragraph{Top-$k$} Our first baseline, top-$k$, ranks all patients being concurrently monitored by their current predicted risk of a critical or emergent ICU transfer within the next 12 hours. Then, it proceeds through this ranking and assigns the highest-intensity feasible action in the order of proactive transfer, CCOT assessment, and Watcher surveillance. For each assignment, the policy respects the remaining intake allowance and the permitted escalations from the patient's current monitoring assignment. 


\paragraph{Threshold-based-top-$k$} Our second baseline, threshold-based-top-$k$, adds an action-specific risk threshold to the allocation rule from top-$k$. That is, a patient is only recommended for a certain action if their predicted risk meets or exceeds the corresponding threshold for that action. We calibrate the thresholds using quantiles of the training period risk score distribution and hold them fixed during evaluation. For each action, the quantile level depends on the combined allowance for that action and all higher-intensity actions relative to the reference ward census. Among qualifying patients, the policy follows the same risk ranking and action order as top-$k$. The thresholds for each action are nested so that a patient who qualifies for transfer can still receive CCOT or Watcher surveillance when the transfer allowance is exhausted. This policy leaves resources unused when no eligible patient meets the corresponding threshold.


\subsection{Evaluation Design}
\label{sec:eval_design}\label{sec:choa_case_study}\label{sec:choa_instantiation}

We evaluate MILOP and the baseline policies over the period from October 1, 2022, through September 30, 2024. The evaluation includes 101,277 ward \emph{episodes} (i.e., individual patient trajectories from the time they enter the ward until they leave) over 731 calendar days, comprising 98,913 non-ICU departures and 2,364 ICU transfers. Among the ICU transfers, 684 are classified as critical and 147 as emergent. At each decision epoch, MILOP and each of the baseline policies generate recommendations using the patient information available at that time.

We evaluate all three policies under the same rolling 24-hour resource budgets: 32 Watcher starts, 25 CCOT starts, and three proactive ICU-transfer recommendations. Each Watcher or CCOT assignment lasts up to 12 hours, ending earlier if the episode leaves the eligible population or the policy recommends a higher-intensity action. We count each new assignment or renewal after expiration as one start for that resource. We charge escalation to the budget for the new action, for example, escalating from Watcher to CCOT uses one CCOT start. Continuing an active assignment adds monitoring time without consuming another start. After a policy recommends proactive transfer, we remove the episode from further consideration by that policy. 

We assess policy performance through coverage, recommendation lead time, and resource use. We call an episode \emph{reached} if the policy recommends Watcher surveillance, CCOT assessment, or proactive transfer strictly before observed ward departure. In our analysis, \emph{coverage} measures the proportion of episodes reached within each outcome category. For action-specific counts, we credit each reached episode to its highest-intensity recommendation before departure. For example, if a patient is recommended both Watcher surveillance and CCOT assessment before departure, we count this episode towards our CCOT assessments. For episodes ending in ICU transfer, we measure \emph{lead time} from the first qualifying recommendation to the observed transfer. We also report the proportion of ICU-transfer episodes that receive a qualifying recommendation at least $T$ hours before observed transfer. We calculate this proportion using all ICU-transfer episodes in the outcome group under analysis, including those without a recommendation. To measure monitoring use, we sum the hours that each policy assigns episodes to Watcher or CCOT monitoring and divide by 24 to obtain patient-days. We report separate counts for Watcher starts, CCOT starts, and proactive ICU-transfer recommendations.


\subsection{Coverage and Usage of Monitoring Resources}
\label{sec:msom-readable-coverage}

We first compare the coverage of each policy and their associated use of monitoring resources, i.e., Watcher surveillance $(a_1)$ or CCOT $(a_2)$. Compared to the baselines, MILOP reaches a greater percentage of episodes ending in critical $(C_2)$ or emergent $(C_3)$ ICU transfer while using less monitoring (Figure~\ref{fig:msom-uniform-M1-coverage-monitoring}). Specifically, MILOP recommends care before observed ward departure for 87.4\% of the 831 critical or emergent episodes, compared with 79.5\% under threshold-based top-$k$ and 70.2\% under top-$k$. Additionally, MILOP's combined usage of Watcher and CCOT resources is 13.5\% lower than threshold-based-top-$k$ and 24.6\% lower than top-$k$. In consideration of all outcomes and resource usage (see Figures~\ref{fig:msom-uniform-S1-outcome-coverage} and~\ref{fig:msom-uniform-M4-paired-effects} in the e-companion), we also find that, compared to the baselines, MILOP uses resources more prudently for episodes that end in departure without ICU transfer (i.e., state $N$). Overall, the lower use of monitoring resources by MILOP primarily stems from lower CCOT use. Relative to the threshold-based-top-$k$ policy, MILOP recommends more Watcher surveillance and slightly more transfer recommendations and  relative to top-$k$, MILOP recommends less of both monitoring resources and the same number of transfer recommendations. The gain therefore involves a different mix of resources, particularly in comparison with the threshold policy. 

\begin{figure}[t!]
\FIGURE
{\includegraphics[width=\textwidth]{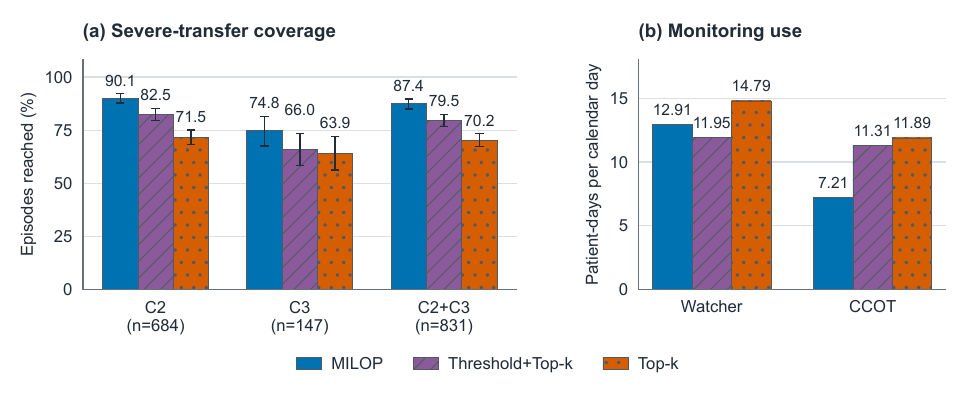}}
{Severe-transfer coverage and monitoring use\label{fig:msom-uniform-M1-coverage-monitoring}}
{Panel (a) reports coverage among critical, emergent, and pooled severe-transfer episodes; denominators are 684, 147, and 831. The plot suggests that MILOP can successfully mark more patients before they start deteriorating under a lower monitoring resource use. Panel (b) reports Watcher and CCOT patient-days per calendar day. Transfer recommendations are a separate resource. Threshold+Top-k denotes the threshold-based top-$k$ policy.}
\end{figure}



\subsection{Lead Time}
\label{sec:msom-readable-timing}

We next examine the timeliness of the coverage by MILOP. Among the episodes reached ending in critical $(C_2)$ or emergent $(C_3)$ ICU transfer, MILOP first recommends some surveillance or transfer intervention $(a_1,a_2,a_3)$ a median of 19.2 hours before observed ICU transfer. 


\begin{figure}[h!]
\FIGURE
{\includegraphics[width=\textwidth]{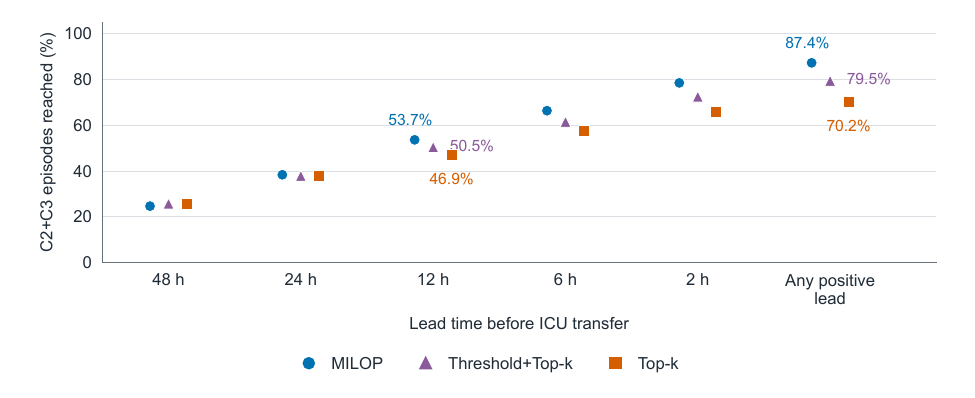}}
{Severe-transfer coverage by warning deadline\label{fig:msom-uniform-M2-warning-deadlines}}
{Each deadline retains all 831 severe episodes, including misses. Coverage requires a first recommendation at least the stated time before observed transfer; the final category requires strictly positive lead. Points sum the reported lead-time buckets without interpolation. Lead time is measured relative to observed transfer, not the first clinical intervention.}
\end{figure}

When analyzing severe transfer coverage stratified by lead time (Figure~\ref{fig:msom-uniform-M2-warning-deadlines}), we find that MILOP reaches 53.7\% of severe episodes at least 12 hours before observed transfer, compared with 50.5\% under threshold-based top-$k$ and 46.9\% under top-$k$. MILOP additionally achieves greater coverage of severe transfers compared to the two baselines across most lead times considered. 
Notably, the difference in coverage across the policies shrinks at longer horizons. Thus, MILOP's advantage over the baselines is concentrated in the hours closer to transfer. However, we remark that MILOP does not necessarily provide more intervention at every horizon.


\begin{figure}[h!]
\FIGURE
{\includegraphics[width=\textwidth]{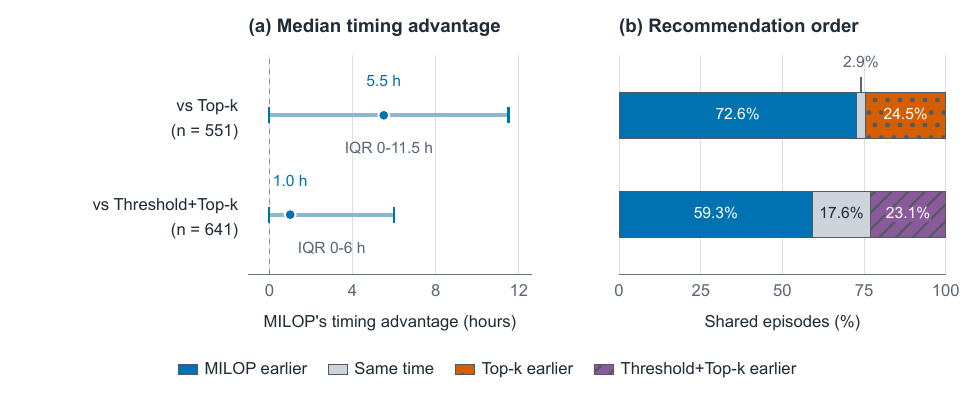}}
{Paired recommendation timing among jointly reached severe episodes\label{fig:msom-uniform-M5-paired-timing-advantage}}
{Comparisons use separate jointly reached severe cohorts: 641 episodes against Threshold+Top-k and 551 against Top-k. Panel (a) shows the median within-episode timing difference (comparator minus MILOP); positive values favor MILOP. Ranges are interquartile ranges, not confidence intervals. Panel (b) partitions the same cohorts by recommendation order.}
\end{figure}

We now additionally present pairwise comparisons of lead time between MILOP and each baseline on the episodes reached by the baseline.Because the policies reach different episodes, we compare their timing using only severe episodes that both MILOP and the respective baseline reach before transfer. Within these separate paired cohorts, MILOP recommends care a median of 1.0 hour earlier than threshold-based top-$k$ and 5.5 hours earlier than top-$k$ (Figure~\ref{fig:msom-uniform-M5-paired-timing-advantage}).These intervals describe an opportunity for earlier assessment and escalation. Section~\ref{sec:msom-ec-timing-details} of the e-companion provides the cohort sizes and timing distributions.



\subsection{Action Sequences and Intensity}
\label{sec:msom-readable-allocation}
To understand how the policies differ in the care they propose, we examine how far a child's care escalates and in what order actions are recommended. Panel (a) of Figure \ref{fig:msom-uniform-M3-staged-recommendations} reports how many of the three escalation responses, Watcher surveillance $(a_1)$, CCOT assessment $(a_2)$, and proactive ICU transfer $(a_3)$, a child received during the ward stay. 
Among children who went on to a critical or emergent transfer and received any response at all, MILOP moved beyond a single response for 68.5\% of them, compared with 56.8\% under threshold-based-top-$k$ and 30.2\% under top-$k$. Under top-$k$, nearly 70\% of these children were acted on only once and then followed under usual care.

\begin{figure}[h!]
\FIGURE
{\includegraphics[width=\textwidth]{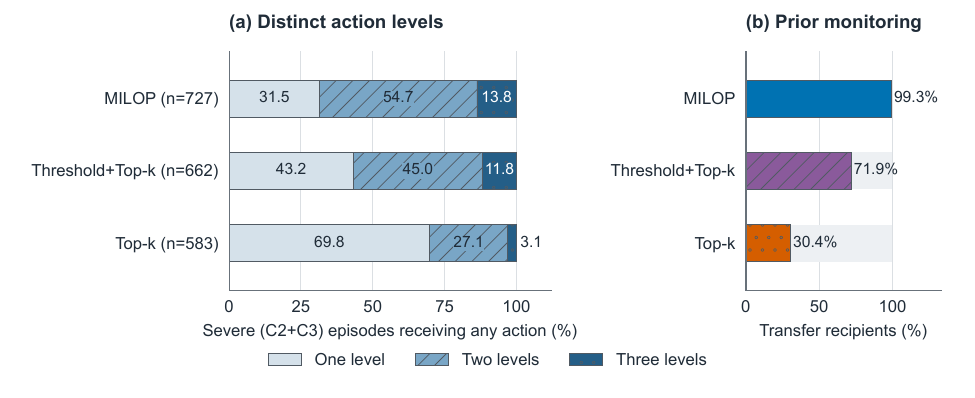}}
{Action levels and monitoring before transfer recommendations\label{fig:msom-uniform-M3-staged-recommendations}}
{Panel (a) counts distinct action levels among critical and emergent episodes receiving assignments; repeated starts at one level do not add levels. This summary does not require every later start to precede departure. Panel (b) uses all strict pre-departure transfer recipients and records whether Watcher or CCOT was recommended earlier. The panels describe different selected populations; supporting counts and timing definitions appear in the e-companion.}
\end{figure}

The order in which monitoring and transfer recommendations are used follows the same pattern. Specifically, MILOP more often begins with Watcher surveillance, whereas the baselines more often start with a higher intensity recommendation. Among all episodes that end in an ICU transfer recommendation ($a_3$) before a recorded ICU transfer, MILOP recommends a Watcher $(a_1)$ and/or CCOT $(a_2)$ action at some point prior to the transfer in 99.3\% of episodes. This proportion is much greater than 71.9\% under threshold-based-top-$k$ and 30.4\% under top-$k$.

Critically, these results show that MILOP's transfer recommendations are usually preceded by an opportunity for Watcher surveillance or CCOT assessment. This sequence is consistent with an ideal staged approach to escalation. That is, MILOP often recommends monitoring actions before recommending ICU transfer, rather than relying on an ICU transfer as the first response. We provide additional analysis that separates first recommendations from the strongest action assigned to each episode in Figure~\ref{fig:msom-uniform-S5-first-strongest-recommendation} of the e-companion.




\subsection{Missed Cases}
\label{sec:msom-readable-misses}

Our final analysis aims to characterize the \emph{missed cases}, i.e., the unreached critical or emergent episodes. Overall, MILOP has a smaller proportion of missed cases compared to either baseline (Table~\ref{tab:msom-missed-summary}). Among the critical and emergent  cases MILOP misses 9.9\% and 25.2\% episodes respectively. Critically, for MILOP, the vast majority (81.0\%) of missed cases never entered either of the highest two risk bins (i.e., states $B_3$ or $B_4$) --- a substantial improvement over the two baselines; 34.1\% and 53.6\% of missed cases for threshold-based-top-$k$ and top-$k$, respectively, had entered the top risk bin $(B_4)$ at some point compared to $0.0\%$ for MILOP. We provide additional details on these missed cases in Figure~\ref{fig:msom-S8-missed-final-bin} and Table~\ref{tab:msom-miss-diagnostics} of the e-companion.



\begin{table}[h!]
\TABLE
{Severity and risk profile of missed cases\label{tab:msom-missed-summary}}
{\begin{tabular*}{\textwidth}{@{\extracolsep{\fill}}lrrr@{}}
\hline\up
Measure & MILOP & Threshold+Top-$k$ & Top-$k$ \down\\
\hline
\multicolumn{4}{@{}l}{\up\textit{Miss rate within each severity group}}\\
Critical episodes missed & 9.9\% & 17.5\% & 28.5\% \\
Emergent episodes missed & 25.2\% & 34.0\% & 36.1\% \down\\
\hline
\multicolumn{4}{@{}l}{\up\textit{Risk profile among missed severe episodes}}\\
Never reached $B_3$ or $B_4$ & 81.0\% & 40.6\% & 23.8\% \down\\
Ever reached $B_4$ & 0.0\% & 34.1\% & 53.6\% \\
\hline
\end{tabular*}}
{Miss rates use all 684 critical or 147 emergent episodes. The final row uses each policy's own missed severe episodes: 105, 170, and 248, respectively. A miss means no qualifying recommendation strictly before observed ward departure. $B_3$ and $B_4$ are the two highest risk bins.}
\end{table}

\section{Managerial Implications}
\label{sec:proposed-managerial}


Our analysis of MILOP and its comparison with two practically relevant baselines leads to several important managerial insights.

\paragraph{An early warning score alone is insufficient. }MILOP and both baselines had access to risk scores for each patient, mimicking the current process that is utilized in many other early warning systems. However, MILOP significantly outperformed both of these two baselines. We hypothesize that a large part of this performance difference is in MILOP's decision-focused loss function during the training process. This loss function allowed our neural network to map both the patient feature vectors and risk scores to index scores and, ultimately, ward escalation decisions that outperformed both baselines in terms of having greater coverage, earlier recommendations, and fewer missed cases. 

\paragraph{Monitoring resources should be used judiciously.} We found that MILOP did not use all of its resources in each decision epoch. This finding was surprising since, in many resource allocation optimization problems, it is inefficient to leave resources unassigned. Upon further analysis, we found that MILOP's neural network was producing index scores that preferred usual care over a monitoring resource (i.e., Watcher surveillance or CCOT assessment) for low-risk patients, even if there were additional resources that could have been used. In contrast, top-$k$ used all of its resources (by design) and threshold-based-top-$k$ used fewer resources than top-$k$ but more resources than MILOP. Despite using fewer resources overall, MILOP outperformed the two baselines. 

\paragraph{Multiple escalation resources should be coordinated jointly.}
In a setting with multiple escalation options, identifying which patients are at greatest risk is only one part of the decision; the hospital must also determine which response is appropriate for each patient while accounting for the distinct capacities and clinical roles of Watcher surveillance, CCOT assessment, and proactive ICU-transfer recommendations. Our results show that extending familiar top-$k$ or threshold-based rules separately to each intervention does not adequately address this coordination problem. For example, a top-$k$ policy may recommend an intensive intervention for a relatively low-risk patient simply because that patient is the highest-risk individual currently on the ward, whereas a threshold-based policy may leave a patient under usual care even when early monitoring would be beneficial because the patient's risk score does not yet exceed a fixed cutoff. MILOP avoids these shortcomings and instead evaluates the competing patient--action assignments jointly within a common allocation problem, allowing monitoring and transfer resources to be used as components of a staged escalation pathway. Indeed, $99.3\%$ of MILOP's proactive ICU-transfer recommendations were preceded by Watcher surveillance or CCOT assessment. Thus, hospitals introducing multiple escalation responses should manage them as a coordinated portfolio rather than as independent interventions layered onto a single early warning score.

\section{Toward Implementation}\label{sec: implementation}

Our team has outlined several steps to move MILOP into care. As described in Section \ref{sec:setting}, teams within Pillar 1 of the CHOA Pediatric Technology Center are already developing both a user interface and technological infrastructure to pair with MILOP. Our results in Section \ref{sec:choa_case_study} show that MILOP's performance \textit{in silico} is promising. Thus, our key next step is to perform a silent evaluation. That is, we will have MILOP implemented in CHOA's EHR system and running in the background. We will then iteratively assess MILOP's prospective performance on real patients, reviewing results with clinicians and technology leaders at CHOA to determine when it is fit for deployment. 
The final implementation will take a phased and iterative approach, wherein we will implement MILOP one department at a time and continue to record its real-time performance with patients. We plan to make adjustments based on feedback from this process until MILOP has been fully deployed.

\Needspace{7\baselineskip}
\section{Conclusion and Future Research}
\label{sec:proposed-conclusion}

Motivated by our partnership with CHOA, this research develops a multi-action RMAB framework for concurrent patient monitoring and ward escalation decisions, along with a novel solution approach called MILOP. In the CHOA setting, this framework serves a broader operational purpose by providing a coordinating layer across escalation resources that previously existed as separate clinical mechanisms rather than as a jointly managed system. MILOP trains a neural network to predict index scores that are then transformed into monitoring and transfer decisions under resource constraints. A key innovation in MILOP is the training process which combines an optimal transport problem and a decision-focused loss function. We show using historical data from CHOA that MILOP outperforms two practically relevant benchmarks; compared to the two benchmarks, MILOP recommends interventions for a greater proportion of critical or emergent transfer episodes with greater lead times and more efficient use of resources. MILOP also generates staged intervention decisions that are well-aligned with clinical intuition. The next phase of this project will focus on implementation as part of CHOA's Pillar 1 program (see Section \ref{sec: implementation}).

This limitations of this research suggest several future research directions. First, our evaluation is retrospective and holds observed patient trajectories and outcomes fixed. Consequently, our analysis assesses the timing and allocation of recommendations and cannot estimate the causal effects of MILOP-guided interventions on subsequent clinical outcomes. Future work should prospectively evaluate the framework after integration into clinical workflows, including how clinicians respond to recommendations and whether coordinated escalation changes patient outcomes. Second, because historical Watcher and CCOT assignment data are unavailable, the effects of these interventions are represented through planning assumptions informed by clinical discussions rather than estimated causal effects. Prospective deployment would create an opportunity to estimate these effects empirically and update the transition model as intervention data accumulate. Third, MILOP depends on the upstream early-warning model to identify patients at risk. Many of the remaining severe episodes missed by MILOP occurred among patients who never entered the highest-risk strata, indicating that improvements in risk estimation may complement improvements in resource allocation. Future research could therefore study joint learning of deterioration risk and downstream allocation policies, rather than treating risk prediction and resource allocation as separate stages.

\bibliographystyle{informs2014}
\bibliography{references,additional_references}

@inproceedings{mate2022field,
  title={Field study in deploying restless multi-armed bandits: Assisting non-profits in improving maternal and child health},
  author={Mate, Aditya and Madaan, Lovish and Taneja, Aparna and Madhiwalla, Neha and Verma, Shresth and Singh, Gargi and Hegde, Aparna and Varakantham, Pradeep and Tambe, Milind},
  booktitle={Proceedings of the AAAI Conference on Artificial Intelligence},
  volume={36},
  pages={12017--12025},
  year={2022}
}

@inproceedings{verma2023restless,
  title={Restless Multi-Armed Bandits for Maternal and Child Health: Results from Decision-Focused Learning.},
  author={Verma, Shresth and Mate, Aditya and Wang, Kai and Madhiwalla, Neha and Hegde, Aparna and Taneja, Aparna and Tambe, Milind},
  booktitle={AAMAS},
  pages={1312--1320},
  year={2023}
}

@inproceedings{wang2023scalable,
  title={Scalable decision-focused learning in restless multi-armed bandits with application to maternal and child health},
  author={Wang, Kai and Verma, Shresth and Mate, Aditya and Shah, Sanket and Taneja, Aparna and Madhiwalla, Neha and Hegde, Aparna and Tambe, Milind},
  booktitle={Proceedings of the AAAI Conference on Artificial Intelligence},
  volume={37},
  pages={12138--12146},
  year={2023}
}

@article{xie2020differentiable,
  title={Differentiable top-k with optimal transport},
  author={Xie, Yujia and Dai, Hanjun and Chen, Minshuo and Dai, Bo and Zhao, Tuo and Zha, Hongyuan and Wei, Wei and Pfister, Tomas},
  journal={Advances in neural information processing systems},
  volume={33},
  pages={20520--20531},
  year={2020}
}

@article{cuturi2019differentiable,
  title={Differentiable ranking and sorting using optimal transport},
  author={Cuturi, Marco and Teboul, Olivier and Vert, Jean-Philippe},
  journal={Advances in neural information processing systems},
  volume={32},
  year={2019}
}

@article{cuturi2013sinkhorn,
  title={Sinkhorn distances: Lightspeed computation of optimal transport},
  author={Cuturi, Marco},
  journal={Advances in neural information processing systems},
  volume={26},
  year={2013}
}

@inproceedings{amos2017optnet,
  title={Optnet: Differentiable optimization as a layer in neural networks},
  author={Amos, Brandon and Kolter, J Zico},
  booktitle={International conference on machine learning},
  pages={136--145},
  year={2017},
  organization={PMLR}
}

@article{agrawal2019differentiable,
  title={Differentiable convex optimization layers},
  author={Agrawal, Akshay and Amos, Brandon and Barratt, Shane and Boyd, Stephen and Diamond, Steven and Kolter, J Zico},
  journal={Advances in neural information processing systems},
  volume={32},
  year={2019}
}

@article{mandi2024decision,
  title={Decision-focused learning: Foundations, state of the art, benchmark and future opportunities},
  author={Mandi, Jayanta and Kotary, James and Berden, Senne and Mulamba, Maxime and Bucarey, Victor and Guns, Tias and Fioretto, Ferdinando},
  journal={Journal of Artificial Intelligence Research},
  volume={80},
  pages={1623--1701},
  year={2024}
}

@inproceedings{wilder2019melding,
  title={Melding the data-decisions pipeline: Decision-focused learning for combinatorial optimization},
  author={Wilder, Bryan and Dilkina, Bistra and Tambe, Milind},
  booktitle={Proceedings of the AAAI Conference on Artificial Intelligence},
  volume={33},
  pages={1658--1665},
  year={2019}
}

@article{donti2017task,
  title={Task-based end-to-end model learning in stochastic optimization},
  author={Donti, Priya and Amos, Brandon and Kolter, J Zico},
  journal={Advances in neural information processing systems},
  volume={30},
  year={2017}
}

@article{amos2018differentiable,
  title={Differentiable mpc for end-to-end planning and control},
  author={Amos, Brandon and Jimenez, Ivan and Sacks, Jacob and Boots, Byron and Kolter, J Zico},
  journal={Advances in neural information processing systems},
  volume={31},
  year={2018}
}

@article{futoma2020popcorn,
  title={Popcorn: Partially observed prediction constrained reinforcement learning},
  author={Futoma, Joseph and Hughes, Michael C and Doshi-Velez, Finale},
  journal={arXiv preprint arXiv:2001.04032},
  year={2020}
}

@book{altman2021constrained,
  title={Constrained Markov decision processes},
  author={Altman, Eitan},
  year={2021},
  publisher={Routledge}
}

@article{adelman2024genetic,
  author = {Adelman, Daniel and Wang, Kanix},
  title = {Frontiers in Operations: Optimal Genetic Testing of Families},
  journal = {Manufacturing \& Service Operations Management},
  year = {2024},
  volume = {26},
  number = {4},
  pages = {1338--1357},
  doi = {10.1287/msom.2023.0057}
}

@article{ayvaci2012budgetary,
  author = {Ayvaci, Mehmet U. S. and Alagoz, Oguzhan and Burnside, Elizabeth S.},
  title = {The Effect of Budgetary Restrictions on Breast Cancer Diagnostic Decisions},
  journal = {Manufacturing \& Service Operations Management},
  year = {2012},
  volume = {14},
  number = {4},
  pages = {600--617},
  doi = {10.1287/msom.1110.0371}
}

@article{gijsbrechts2022inventory,
  author = {Gijsbrechts, Joren and Boute, Robert N. and Van Mieghem, Jan A. and Zhang, Dennis J.},
  title = {Can Deep Reinforcement Learning Improve Inventory Management? {P}erformance on Lost Sales, Dual-Sourcing, and Multi-Echelon Problems},
  journal = {Manufacturing \& Service Operations Management},
  year = {2022},
  volume = {24},
  number = {3},
  pages = {1349--1368},
  doi = {10.1287/msom.2021.1064}
}

@article{chan2023milk,
  title={Got (Optimal) Milk? Pooling Donations in Human Milk Banks with Machine Learning and Optimization},
  author={Chan, Timothy C. Y. and Mahmood, Rafid and O'Connor, Deborah L. and Stone, Debbie and Unger, Sharon and Wong, Rachel K. and Zhu, Ian Yihang},
  journal={Manufacturing \& Service Operations Management},
  year={2023},
  note={Published online November 17, 2023},
  doi={10.1287/msom.2022.0455}
}

@misc{ye2025contextual,
  title={Contextual Stochastic Optimization for Omnichannel Multi-Courier Order Fulfillment Under Delivery Time Uncertainty},
  author={Ye, Tinghan and Cheng, Sikai and Hijazi, Amira and Van Hentenryck, Pascal},
  year={2025},
  note={arXiv:2409.06918v2},
  url={https://arxiv.org/abs/2409.06918v2}
}

@article{bonafide2014critical,
  title={Impact of Rapid Response System Implementation on Critical Deterioration Events in Children},
  author={Bonafide, Christopher P. and Localio, A. Russell and Roberts, Kathryn E. and Nadkarni, Vinay M. and Weirich, Christine M. and Keren, Ron},
  journal={JAMA Pediatrics}, year={2014}, volume={168}, number={1}, pages={25--33},
  doi={10.1001/jamapediatrics.2013.3266}
}

@article{mehta2022late,
  title={Leveraging {EHR} Data to Evaluate the Association of Late Recognition of Deterioration With Outcomes},
  author={Mehta, Sanjiv D. and Muthu, Naveen and Yehya, Nadir and Galligan, Meghan and Porter, Ezra and McGowan, Nancy and Papili, Kelly and Favatella, Dana and Liu, Hongyan and Griffis, Heather and Bonafide, Christopher P. and Sutton, Robert M.},
  journal={Hospital Pediatrics}, year={2022}, volume={12}, number={5}, pages={447--460},
  doi={10.1542/hpeds.2021-006363}
}

@misc{gatech2026ipat,
  author={{Georgia Institute of Technology}},
  title={{IPaT} Research Scientists Supporting Pediatric Research},
  year={2026},
  howpublished={Research News Center, February 27},
  url={https://news.research.gatech.edu/2026/02/27/ipat-research-scientists-supporting-pediatric-research},
  note={Accessed September 16, 2026}
}

@misc{gatech2026accord,
  author={{Georgia Institute of Technology}},
  title={{AI} Tool Aims to Catch Critically Ill Children},
  year={2026},
  howpublished={Research News Center, September 15},
  url={https://news.research.gatech.edu/2026/09/15/ai-tool-aims-catch-critically-ill-children},
  note={Accessed September 16, 2026}
}

@article{anonymous_pediatric_icu_transfer,
  title={Early Prediction of Pediatric {ICU} Transfer: A Comparative Study of Tabular and Time Series Machine Learning Models},
  author={{Anonymous}},
  year = {2026},
  journal={INFORMS Journal on Data Science},
  note={Manuscript under review},
  publisher={INFORMS}
}

@article{hu2018examination,
  title={An examination of early transfers to the ICU based on a physiologic risk score},
  author={Hu, Wenqi and Chan, Carri W and Zubizarreta, Jos{\'e} R and Escobar, Gabriel J},
  journal={Manufacturing \& Service Operations Management},
  volume={20},
  number={3},
  pages={531--549},
  year={2018},
  publisher={INFORMS}
}

@article{lambert2017paediatric,
  title={Paediatric early warning systems for detecting and responding to clinical deterioration in children: a systematic review},
  author={Lambert, Veronica and Matthews, Anne and MacDonell, Rachel and Fitzsimons, John},
  journal={BMJ open},
  volume={7},
  number={3},
  pages={e014497},
  year={2017},
  publisher={British Medical Journal Publishing Group}
}

@article{parshuram2018effect,
  title={Effect of a pediatric early warning system on all-cause mortality in hospitalized pediatric patients: the EPOCH randomized clinical trial},
  author={Parshuram, Christopher S and Dryden-Palmer, Karen and Farrell, Catherine and Gottesman, Ronald and Gray, Martin and Hutchison, James S and Helfaer, Mark and Hunt, Elizabeth A and Joffe, Ari R and Lacroix, Jacques and others},
  journal={Jama},
  volume={319},
  number={10},
  pages={1002--1012},
  year={2018}
}

@article{chong2022paediatric,
  title={Do paediatric early warning systems reduce mortality and critical deterioration events among children? A systematic review and meta-analysis},
  author={Chong, Shu-Ling and Goh, Mark Sen Liang and Ong, Gene Yong-Kwang and Acworth, Jason and Sultana, Rehena and Yao, Sarah Hui Wen and Ng, Kee Chong and Scholefield, Barney and Aickin, Richard and Maconochie, Ian and others},
  journal={Resuscitation plus},
  volume={11},
  pages={100262},
  year={2022},
  publisher={Elsevier}
}

@article{evans2021implementing,
  title={Implementing a watcher program to improve timeliness of recognition of deterioration in hospitalized children},
  author={Evans, Stephanie and Green, Angela and Roberson, Angela and Webb, Tammy and Edwards, Christopher},
  journal={Journal of Pediatric Nursing},
  volume={61},
  pages={151--156},
  year={2021},
  publisher={Elsevier}
}

@article{buist2004association,
  title={Association between clinically abnormal observations and subsequent in-hospital mortality: a prospective study},
  author={Buist, Michael and Bernard, Stephen and Nguyen, Tuan V and Moore, Gaye and Anderson, Jeremy},
  journal={Resuscitation},
  volume={62},
  number={2},
  pages={137--141},
  year={2004},
  publisher={Elsevier}
}

@article{helen1999unexpected,
  title={Unexpected Deaths and Referrals to Intensive Care of Patients on General Wards--Are Some Cases Potentially Avoidable?},
  author={Helen McGloin, RGN and Adam, Sheila K and Singer, Mervyn and others},
  journal={Journal of the Royal College of Physicians of London},
  volume={33},
  number={3},
  pages={255},
  year={1999}
}

@misc{xu_reinforcement_2025,
	title = {Reinforcement learning with combinatorial actions for coupled restless bandits},
	copyright = {arXiv.org perpetual, non-exclusive license},
	url = {https://arxiv.org/abs/2503.01919},
	doi = {10.48550/ARXIV.2503.01919},
	urldate = {2025-10-24},
	author = {Xu, Lily and Wilder, Bryan and Khalil, Elias B. and Tambe, Milind},
	year = {2025},
	note = {Publisher: arXiv
Version Number: 2},
}

@misc{zamir_deep_2024,
	title = {Deep {Index} {Policy} for {Multi}-{Resource} {Restless} {Matching} {Bandit} and {Its} {Application} in {Multi}-{Channel} {Scheduling}},
	url = {http://arxiv.org/abs/2408.07205},
	doi = {10.48550/arXiv.2408.07205},
	urldate = {2025-10-19},
	publisher = {arXiv},
	author = {Zamir, Nida and Hou, I.-Hong},
	month = aug,
	year = {2024},
	note = {arXiv:2408.07205 [cs]},
}

@article{hodge_asymptotic_2015,
	title = {On the asymptotic optimality of greedy index heuristics for multi-action restless bandits},
	volume = {47},
	copyright = {https://www.cambridge.org/core/terms},
	issn = {0001-8678, 1475-6064},
	url = {https://www.cambridge.org/core/product/identifier/S0001867800048771/type/journal_article},
	doi = {10.1239/aap/1444308876},
	language = {en},
	number = {3},
	urldate = {2025-10-19},
	journal = {Advances in Applied Probability},
	author = {Hodge, D. J. and Glazebrook, K. D.},
	month = sep,
	year = {2015},
	pages = {652--667},
}

@misc{goldsztajn_asymptotically_2024,
	title = {Asymptotically {Optimal} {Policies} for {Weakly} {Coupled} {Markov} {Decision} {Processes}},
	url = {http://arxiv.org/abs/2406.04751},
	doi = {10.48550/arXiv.2406.04751},
	urldate = {2025-10-19},
	publisher = {arXiv},
	author = {Goldsztajn, Diego and Avrachenkov, Konstantin},
	month = dec,
	year = {2024},
	note = {arXiv:2406.04751 [math]},
}

@article{verloop_asymptotically_2016,
	title = {Asymptotically optimal priority policies for indexable and nonindexable restless bandits},
	volume = {26},
	issn = {1050-5164},
	url = {http://arxiv.org/abs/1609.00563},
	doi = {10.1214/15-AAP1137},
	number = {4},
	urldate = {2025-10-19},
	journal = {The Annals of Applied Probability},
	author = {Verloop, I. M.},
	month = aug,
	year = {2016},
	note = {arXiv:1609.00563 [math]},
}

@article{weber_index_1990,
	title = {On an index policy for restless bandits},
	volume = {27},
	number = {3},
	journal = {Journal of Applied Probability},
	author = {Weber, Richard and Weiss, Gerardo},
	year = {1990},
	pages = {637--648},
}

@article{papadimitriou_complexity_1999,
	title = {The complexity of optimal queueing network control},
	volume = {24},
	number = {2},
	journal = {Mathematics of Operations Research},
	author = {Papadimitriou, Christos H and Tsitsiklis, John N},
	year = {1999},
	pages = {293--305},
}

@article{whittle_restless_1988,
	title = {Restless bandits: {Activity} allocation in a changing world},
	volume = {25},
	journal = {Journal of Applied Probability},
	author = {Whittle, Peter},
	year = {1988},
	pages = {287--298},
}

@article{grand-clement_robustness_2023,
	title = {Robustness of {Proactive} {Intensive} {Care} {Unit} {Transfer} {Policies}},
	volume = {71},
	issn = {0030-364X, 1526-5463},
	url = {https://pubsonline.informs.org/doi/10.1287/opre.2022.2403},
	doi = {10.1287/opre.2022.2403},
	language = {en},
	number = {5},
	urldate = {2025-04-28},
	journal = {Operations Research},
	author = {Grand-Clément, Julien and Chan, Carri W. and Goyal, Vineet and Escobar, Gabriel},
	month = sep,
	year = {2023},
	pages = {1653--1688},
}

\ECSwitch
\ECHead{E-Companion}
\section{CHOA Model Specification}
\label{app:choa_details}

This section describes the risk strata, transition kernels, reward parameters, and resource conventions used in the CHOA application. Usual-care dynamics are estimated from historical ward records. Additional intervention effects and costs are planning assumptions informed by clinical discussions. The stationary planning model and retrospective replay use different resource conventions, which are distinguished below.

\subsection{Risk-Bin Definition}
\label{ec:risk_bins}
The five risk strata use thresholds fitted to the training-period score distribution and then held fixed. The rounded thresholds for the date-based split are 0.000398, 0.001535, 0.007196, and 0.027237.

\subsection{Transition Kernel Construction}
\label{ec:transition_construction}
\paragraph{Usual-care dynamics.} For transition probability $P^a(s,s')$, let $p_i(c)$ denote the usual-care exit probability from $B_i$ to $c\in\mathcal E$, estimated per epoch of eligible ward exposure, and let $Q_{ij}$ describe movement between risk strata conditional on remaining on the ward. Separating exits from within-ward movement gives the usual-care kernel
\begin{align}
P^{a_0}(B_i,c)&=p_i(c),\quad c\in\mathcal{E},\label{eq:ec_usual_exit}\\
P^{a_0}(B_i,B_j)&=\left[1-\sum_{c\in\mathcal{E}}p_i(c)\right]Q_{ij}.
\label{eq:usual_kernel}
\end{align}
We estimate $Q$ on the 30-minute decision grid, counting self-transitions during carried-forward exposure and movement to the next observed stratum. Both movement and exits therefore use exposure measured in half-hour epochs. The exposure credited to an observation is capped at 24 hours, but the estimator retains the next recorded movement or exit label even when it occurs after that cap. It thus attributes that later event to the last observed stratum rather than censoring it when the score becomes stale. This differs from the replay's eligibility rule, which removes patients during stale-score gaps.

\paragraph{Watcher and CCOT scenarios.} Without historical Watcher and CCOT assignment labels, we represent their effects through scenario parameters informed by clinical discussions. The reference remaining-risk factors for critical and emergent transfers are $(\rho_{12},\rho_{13})=(0.75,0.89)$ for Watcher and $(\rho_{22},\rho_{23})=(0.60,0.67)$ for CCOT. For action $a_\ell$, $\ell=1,2$, the reduction in severe-transfer probability and its redistribution are
\begin{align}
d_{i\ell}&=(1-\rho_{\ell2})p_i(C_2)+(1-\rho_{\ell3})p_i(C_3),\\
P^{a_\ell}(B_i,C_k)&=\rho_{\ell k}p_i(C_k),\qquad k=2,3,\\
P^{a_\ell}(B_i,C_0)&=p_i(C_0)+\eta_i d_{i\ell},\\
P^{a_\ell}(B_i,C_1)&=p_i(C_1)+(1-\eta_i)d_{i\ell},
\label{eq:monitor_kernel}
\end{align}
where $\eta_i=p_i(C_0)/[p_i(C_0)+p_i(C_1)]$, with $\eta_i=1$ when the denominator is zero. Redistributing the reduction in $C_2$ and $C_3$ probability to $C_0$ and $C_1$ preserves the total ICU-transfer probability, while leaving non-ICU exits and movement between ward strata unchanged. The scenario therefore represents earlier care that changes transfer severity without necessarily preventing ICU admission.

\paragraph{Proactive-transfer scenario.} Under $a_3$, the next state is an ICU-transfer outcome whose distribution is constructed from the observed outcomes $H_i(c)$ among patients transferred from stratum $i$. An assumed handover interval $h=2$ hours determines the remaining probability of a critical or emergent outcome. With $\Delta=0.5$ hours, this probability is
\begin{equation}
\lambda_i=1-\left[1-p_i(C_2)-p_i(C_3)\right]^{h/\Delta}.
\label{eq:handover}
\end{equation}
We distribute probability $1-\lambda_i$ across $C_0,C_1$ according to their conditional $H_i$ proportions, and probability $\lambda_i$ across $C_2,C_3$ in the same way. Empty groups default to $C_0$ and $C_2$, respectively, while a stratum with fewer than 20 historical ICU transfers borrows the outcome distribution from the nearest adequately supported stratum. The handover rule in \eqref{eq:handover} assumes a constant stratum-specific hazard and adjusts the outcome distribution without modeling a transfer queue. Because $H_i$ reflects patients selected for transfer by clinicians, applying it to a new recommendation is a scenario assumption.

\paragraph{Slot refill.} The admission distribution $\nu$ uses each episode's first scored stratum. For each $e\in\mathcal E$, the refill transition is $P^{a_0}(e,B_j)=\nu_j$, and $a_0$ is the only admissible action at an exit.

The monitoring effects and handover interval jointly shape the value of escalation in the planning model. The results below evaluate recommendations on fixed historical trajectories; they do not estimate the clinical effects assumed in these kernels.

\subsection{Reward Parameters}
\label{ec:reward_parameters}
Table~\ref{tab:reward_params} combines exit penalties based on clinician-reported mortality differences with ward and intervention costs that represent additional burdens. These components define a common reward scale for planning, without giving the total reward a mortality-probability or monetary interpretation.

Let $u_i$ be the cost per 30-minute epoch in ward state $B_i$, let $c_1$ and $c_2$ be monitoring costs per patient-day, and let $c_3$ be the cost per transfer recommendation. The ward reward is
\begin{equation}
r(B_i,a)=-u_i-\frac{c_1}{48}\mathbf{1}\{a=a_1\}
-\frac{c_2}{48}\mathbf{1}\{a=a_2\}
-c_3\mathbf{1}\{a=a_3\}.
\label{eq:ec_ward_reward}
\end{equation}
Usual care has no additional action charge. Exit rewards are incurred once, during the exit-state epoch. The factors of 48 convert daily monitoring costs to the model's decision interval; the transfer cost is an event charge.

\begin{table}[ht]
\TABLE
{CHOA reward parameters. Exit penalties are calibrated to reported mortality differences relative to $C_1$. Ward and action costs are additional scenario weights.\label{tab:reward_params}}
{\begin{tabular}{>{\raggedright\arraybackslash}p{0.32\linewidth}>{\raggedright\arraybackslash}p{0.17\linewidth}>{\raggedright\arraybackslash}p{0.41\linewidth}}
\hline\up
Parameter & Value & Meaning \down\\
\hline\up
$r(N,a_0)=r(C_0,a_0)=r(C_1,a_0)$ & $0$ & Reference outcome \\
$r(C_2,a_0)$ & $-4$ & Critical transfer: $5\%-1\%=4$ percentage points \\
$r(C_3,a_0)$ & $-11$ & Emergent transfer: $12\%-1\%=11$ percentage points \\
$u_0,u_1,u_2$ (ward costs) & $0$ / epoch & No ward-state penalty in $B_0,B_1,B_2$ \\
$u_3$ (ward cost, $B_3$) & $0.10$ / epoch & $4.8$/patient-day; charged under every action \\
$u_4$ (ward cost, $B_4$) & $0.20$ / epoch & $9.6$/patient-day; charged under every action \\
$c_1$ (Watcher cost) & $0.008$ / patient-day & Divided by 48 in Equation~\eqref{eq:ec_ward_reward} \\
$c_2$ (CCOT cost) & $0.030$ / patient-day & Divided by 48 in Equation~\eqref{eq:ec_ward_reward} \\
$c_3$ (transfer-event cost) & $0.50$ / event & Assumed scenario weight \down\\
\hline
\end{tabular}}
{}
\end{table}

\paragraph{Clinician-reported CHOA mortality rates.}\label{ec:mortality_inputs}
CHOA clinicians report in-hospital mortality, measured as death before discharge, of 1\% after $C_1$ transfers, 5\% after $C_2$ transfers, and 12\% after $C_3$ transfers. The zero rewards for $C_0$ and $N$ are utility reference values, not estimates of zero mortality. The reported differences describe associations with transfer category and cannot identify the mortality effect of a policy that changes transfer timing.

Ward costs $u_3$ and $u_4$ accrue under every action, including during an epoch in which transfer is recommended, and can favor earlier departure from high-risk strata independently of the exit penalties. Their effect on the preferred policy must therefore be considered alongside the intervention assumptions and the assumed transfer cost $c_3$.

\subsection{Planning Capacities and Replay Quotas}
\label{ec:capacity_execution}
\label{ec:capacity_values}
The reference planning model uses $M=325$ patient slots, with $b_1=32$ concurrent Watcher assignments and $b_2=25$ concurrent CCOT assignments. Its average proactive-transfer allowance is $f_3=3.25$ recommendations per day, or $\beta_3=f_3/48\simeq0.0677$ per epoch. In a single-slot occupancy LP normalized to total mass one, the corresponding resource constraints are
\begin{align}
\sum_{i=0}^{4}\omega(B_i,a_\ell)&\leq b_\ell/M,
\qquad \ell=1,2,\\
\sum_{i=0}^{4}\omega(B_i,a_3)&\leq f_3/(48M).
\label{eq:ec_planning_budgets}
\end{align}
The fractional transfer rate is an average planning allowance, not an upper bound on an integer transfer count at each epoch. It does not measure available ICU beds or guarantee clinical acceptance of a transfer.

The retrospective replay instead applies common rolling 24-hour intake quotas of 32 Watcher starts, 25 CCOT starts, and three proactive-transfer recommendations. A new assignment, renewal, or escalation consumes the corresponding start allowance; continuation of an active assignment contributes monitoring time without consuming another start. Watcher and CCOT listings last up to 12 hours and end earlier when an episode leaves the eligible population or receives a higher-intensity recommendation. After a transfer recommendation, the episode leaves that policy's subsequent decision population.

A start quota and a concurrent-assignment limit measure different demands. The replay's 12-hour listing window is an allocation convention and should not be interpreted as an estimated duration of delivered clinical care. The replay comparisons therefore evaluate the policies under the stated intake rules; they do not establish performance under the planning model's concurrent capacities or clinical effects under its intervention kernels.

\subsection{Policy Network and Features}
\label{ec:network_config}
This subsection specifies the network inputs, architecture, and training settings, so that the scoring stage can be reimplemented independently of the allocation layer.

\paragraph{Patient features.} With five risk strata, the feature vector $\bm{x}_m$ has nine entries: five stratum indicators and four continuous measures summarizing the current score, elapsed stay, and recent score history. Table~\ref{tab:policy_features} states each entry. Every measure uses only information recorded at or before the decision epoch, and score-history entries are grouped by episode, so no quantity is carried across patients. Between observations, the recorded score, preceding score, and score difference remain fixed at their last observed values, while elapsed stay is re-evaluated at the decision epoch. The score difference is a difference between two observations rather than a derivative. When observations are sparse, the earlier observation nearest one hour back can sit several hours back, and no cap is imposed on that separation.

\begin{table}[ht]
\TABLE
{Policy network inputs. Five stratum indicators and four continuous measures give nine entries. All are available at the decision epoch $t$. Let $o(t)$ denote the most recent scored observation at or before $t$ for the patient's episode.\label{tab:policy_features}}
{\begin{tabular}{>{\raggedright\arraybackslash}p{0.20\linewidth}>{\raggedright\arraybackslash}p{0.10\linewidth}>{\raggedright\arraybackslash}p{0.58\linewidth}}
\hline\up
Entry & Count & Definition \down\\
\hline\up
Stratum indicators & 5 & One-hot encoding of the ward state $S_{mt}\in\{B_0,\ldots,B_4\}$ implied by the score at $o(t)$ \\
Current score & 1 & Early warning score at $o(t)$: the estimated probability of a critical or emergent ICU transfer within 12 hours \\
Elapsed stay & 1 & $\min\{\Delta_m(t)/14,\,1\}$, where $\Delta_m(t)$ is the time in days from the episode's first scored observation to $t$; re-evaluated at every decision epoch, including epochs on which no new score was recorded \\
Preceding score & 1 & Score at the observation immediately before $o(t)$ in the same episode; equals the current score at the episode's first observation \\
Score difference & 1 & Current score minus the score at the earlier observation of the same episode whose timestamp is nearest one hour before $o(t)$; ties resolve to the older observation and the value is zero at the first observation \down\\
\hline
\end{tabular}}
{}
\end{table}

\paragraph{Architecture.} The reference network maps the nine inputs through a stem of width 128 comprising a linear map, layer normalization, a rectified linear unit, and dropout, then four residual blocks, then a linear head with four outputs and no activation. Each residual block applies a linear map to bottleneck width 32, layer normalization, a rectified linear unit, a linear map back to width 128, layer normalization, a rectified linear unit, and dropout, added to its input. Dropout is $0.2$ in the stem and in every block. The network has $36{,}740$ trainable parameters. The same parameters score every patient, so the size of the network does not depend on the number of patients present.

\paragraph{Training.} Table~\ref{tab:training_settings} lists the reference settings. Optimization uses Adam with a fixed learning rate and gradient-norm clipping; no learning-rate schedule or weight decay is applied. Each training step draws one census snapshot, scores all eligible patients, forms the transport plan, and backpropagates the loss in \eqref{eq: training loss} through the Sinkhorn iterations. Snapshot order is permuted once from the run seed, and the same seed initializes the network, so a run is reproducible given identical inputs. Snapshots with fewer than two eligible patients are skipped, since the allocation is then degenerate. Validation snapshots are drawn from a later, disjoint window, are held fixed across epochs, and are used only for checkpoint selection. These are reference implementation settings rather than independently verified settings of the reported checkpoint.

\begin{table}[ht]
\TABLE
{Reference training and transport settings.\label{tab:training_settings}}
{\begin{tabular}{>{\raggedright\arraybackslash}p{0.34\linewidth}>{\raggedright\arraybackslash}p{0.16\linewidth}>{\raggedright\arraybackslash}p{0.38\linewidth}}
\hline\up
Setting & Value & Note \down\\
\hline\up
Optimizer & Adam & No schedule, no weight decay \\
Learning rate & $10^{-4}$ & Fixed \\
Gradient-norm clip & $1.0$ & Applied before each step \\
Training epochs & $20$ & Full passes over sampled snapshots \\
Training snapshots & $\leq 500$ & Sampled decision epochs \\
Validation snapshots & $\leq 200$ & Fixed across epochs \\
Minimum census & $2$ patients & Thinner snapshots skipped \\
Validation interval & $5$ epochs & From epoch $0$, plus the final epoch \\
Early stopping & $4$ checks & Consecutive checks without improvement \\
Transport temperature $\epsilon$ & $0.05$ & Applied to the normalized cost \\
Sinkhorn iterations & $\leq 200$ & Stops early at tolerance $10^{-9}$ \\
Dropout & $0.2$ & Stem and residual blocks \down\\
\hline
\end{tabular}}
{}
\end{table}

\paragraph{Transport numerics.} Two conditioning steps precede the Sinkhorn iterations and determine what the temperature $\epsilon$ means. First, each patient's four scores are centered by their own mean. Adding a constant to one patient's scores does not change that patient's preference ordering, and since each patient row in \eqref{eq: OT problem new} carries fixed mass, it shifts every feasible plan's objective by the same constant. Without this centering, one patient's score offset would change the effective temperature for the whole census. Second, the resulting cost matrix is divided by its largest absolute entry before $\epsilon$ is applied, so the temperature is defined against a cost on a unit scale rather than against the magnitude the network's outputs happen to reach during training. Without this rescaling, a large score scale saturates the plan into a hard assignment whose gradient vanishes. The iterations run in the log domain for numerical stability, and the converged plan is restored to unit patient row mass by a single common factor rather than by normalizing each row, since per-row normalization would break the column capacity bound whenever the iteration has not fully converged.

A finite iteration budget leaves the marginals imperfectly matched, and we report that error rather than assuming convergence. At the reference setting of 200 iterations and a census near the reference value of $M=325$, patient row-sum error is near $10^{-6}$ in most snapshots, with a worst tested case near $5\times10^{-3}$. At 100 iterations the error is near $10^{-2}$, which is not sufficient at this census. Row-sum error and budget excess are measured before the rows are normalized for the loss, so that a convergence failure cannot be concealed by the normalization.

\paragraph{Target construction.} In the reference configuration the target $\pi^{\omega^*}(a\mid S_{mt})$ is a step function of the ward state, so all patients sharing a stratum receive identical targets. Because the Kullback--Leibler objective is convex, it then drives same-stratum patients toward identical scores, and the three history measures cannot change the target within a stratum. Under this setting they are carried so that the input contract is stable as the target construction changes, not because they are informative, and residual within-stratum score variation is close to zero. An optional variant interpolates between neighboring target rows using training-stratum median scores as anchors, which gives the current score a within-stratum gradient, though the resulting targets are no longer themselves an occupancy-LP solution.

\subsection{Solver Settings and Computation Times}
\label{ec:computation}
Three optimization problems are solved at different points and on different schedules, so their costs are not comparable.

\paragraph{Occupancy LP.} Problem \eqref{eq:choa_lp} is solved once, offline, on a single chain with ten states and four actions. Every patient in a stratum shares a kernel and the budgets couple patients only through sums, so the $M$-patient program is this one program symmetrized, and its size does not grow with the census. It is solved with a sparse linear programming solver, and its optimal value is retained as the planning bound. This step is negligible relative to training and does not recur at execution.

\paragraph{Training step.} Each step scores one census snapshot, runs up to 200 log-domain Sinkhorn iterations, and backpropagates through them. At a census of 325 patients this costs approximately 30 milliseconds per snapshot on a single processor core, forward and backward combined. The dominant term is the iteration count rather than the network, which has fewer than $4\times10^4$ parameters.

\paragraph{Execution.} At each decision epoch, scoring the census and solving \eqref{eq: knapsack problem} together take approximately 6 milliseconds at a census of 325, with a 95th percentile near 7 milliseconds. The assignment is posed as a sparse linear program with one equality row per patient and one capacity row per budgeted action, giving $4M$ variables and $M+3$ rows. Two reductions keep this stable at the observed census. The candidate set is first reduced to three times the total capacity in rows, which is exact by an exchange argument, since no optimal solution can assign a scarce action to a patient outside that set. 

A decision epoch is 30 minutes and the reference census is 325 patients, so execution uses a small fraction of the available interval, and the binding constraint on deployment is data availability rather than solve time. These timings characterize the scale of the computation rather than the performance of a deployed system.

\section{Additional Policy Comparisons}
\label{sec:msom-illustrated-ec}

The replay covers 101,277 ward episodes from October 1, 2022, through September 30, 2024: 98,913 non-ICU departures and 2,364 ICU transfers. The latter comprise 1,472 $C_0$, 61 $C_1$, 684 critical ($C_2$), and 147 emergent ($C_3$) episodes. Severe episodes are the 831 $C_2+C_3$ transfers. Unless stated otherwise, policy comparisons are ordered MILOP, Threshold+Top-$k$, and Top-$k$.

An episode is reached if it receives a qualifying Watcher, CCOT, or transfer recommendation strictly before observed ward departure. This definition records whether a recommendation occurred; it does not require monitoring to remain active until transfer. Episode coverage, assignment starts, monitoring patient-time, and issued transfer recommendations are distinct quantities. The analyses below retain their corresponding denominators.

\subsection{Paired Coverage Differences}
\label{sec:msom-ec-coverage-details}
MILOP reaches 726 of 831 severe episodes, compared with 661 under Threshold+Top-$k$ and 583 under Top-$k$. The paired coverage differences are 7.8 percentage points (95\% episode-bootstrap interval: 5.4--10.2) and 17.2 points (14.0--20.3), respectively. Emergent-transfer comparisons use point estimates.

Figure~\ref{fig:msom-uniform-M4-paired-effects} supplements the main manuscript's coverage comparison. It reports paired differences for critical and pooled severe episodes, together with selection differences among episodes without observed ICU transfer. The signs are chosen so that positive values indicate higher severe coverage or lower $N$ selection under MILOP.

The paired episode bootstrap resamples ward episodes while retaining the policies' saved recommendation indicators and the historical outcome for each sampled episode. Each draw therefore compares the policies on the same resampled episodes. The allocation processes are not rerun, so the intervals do not capture changes in competition for resources under a different population. They also omit repeated-stay clustering, calendar-time dependence, policy-selection and model-fitting uncertainty, and multiplicity adjustment. The reported severe-coverage intervals exclude zero under this resampling scheme; they do not establish the direction or magnitude of uncertainty under a broader design.

\begin{figure}[!htbp]
\FIGURE
{\includegraphics[width=\textwidth]{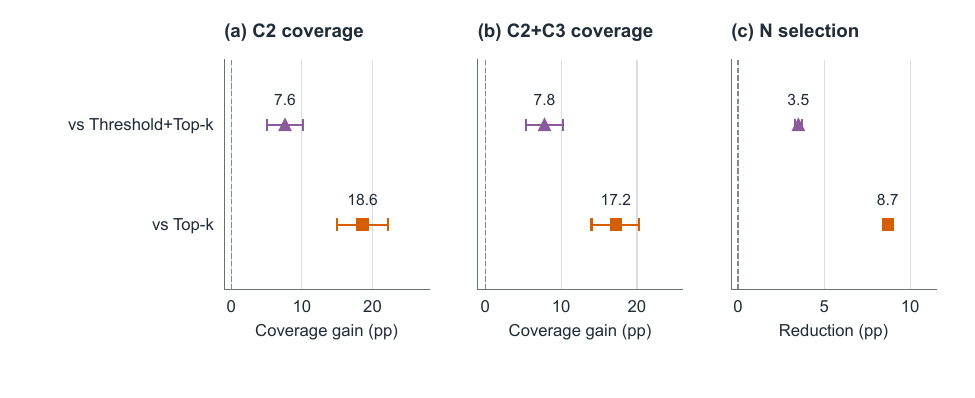}}
{Paired differences in coverage and selection\label{fig:msom-uniform-M4-paired-effects}}
{Whiskers are paired 95\% episode-bootstrap intervals with policies and allocations fixed. Coverage gain is calculated as MILOP minus comparator; $N$ selection reduction is calculated as MILOP minus comparator.} 
\end{figure}

\subsection{Coverage and Action Attribution by Outcome}
Figure~\ref{fig:msom-uniform-S1-outcome-coverage} retains every episode in its observed outcome group and assigns each reached episode to its strongest qualifying recommendation. MILOP reaches 1,024 $C_0$ and 50 $C_1$ episodes, compared with 947 and 46 under Threshold+Top-$k$ and 828 and 37 under Top-$k$. The $C_1$ denominator is only 61 episodes. The $N$ group describes selection without observed ICU transfer and does not adjudicate unnecessary care.

Attribution requires a recommendation strictly before observed ward departure and follows the ordering Watcher, CCOT, then transfer. An episode that receives Watcher followed by CCOT is counted once under CCOT; if it subsequently receives a qualifying transfer recommendation, it is counted once under transfer. Episodes without a qualifying recommendation form the no-action category. These mutually exclusive categories sum to each outcome total. They differ from counts of every action received or every new assignment start.

\begin{figure}[!htbp]
\FIGURE
{\includegraphics[width=\textwidth]{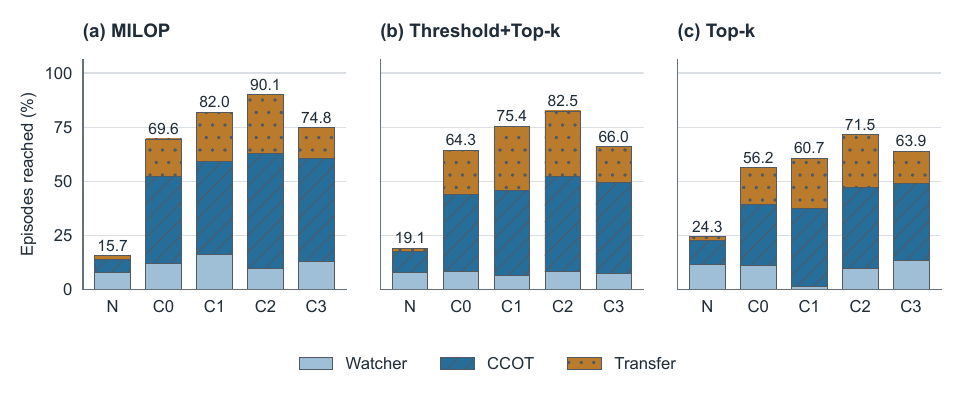}}
{Coverage across all outcome groups stratified by recommended action\label{fig:msom-uniform-S1-outcome-coverage}}
{Episodes are credited once to their strongest recommendation strictly before departure; bar labels indicate coverage. } 
\end{figure}

\subsection{First and Strongest Recommendations}
\label{sec:msom-ec-pathway-details}
Figure~\ref{fig:msom-uniform-S5-first-strongest-recommendation} compares the first and strongest qualifying actions among each policy's reached severe episodes. These are marginal distributions within policy-specific populations; the bars alone do not reconstruct complete episode-level transitions. The main-text transfer-pathway comparison separately records whether monitoring preceded a transfer recommendation.

Across all strict pre-departure transfer recipients, prior monitoring occurs in 2,176 of 2,192 MILOP episodes, 1,534 of 2,135 Threshold+Top-$k$ episodes, and 667 of 2,192 Top-$k$ episodes. These are the denominators for the main-text prior-monitoring percentages. Transfer is the first recommendation in the remaining 16, 601, and 1,525 episodes. Among recipients with prior monitoring, median time from first monitoring to transfer recommendation is 10.5, 11.0, and 48.0 hours, respectively. These intervals concern separate selected populations.

\begin{figure}[!htbp]
\FIGURE
{\includegraphics[width=\textwidth]{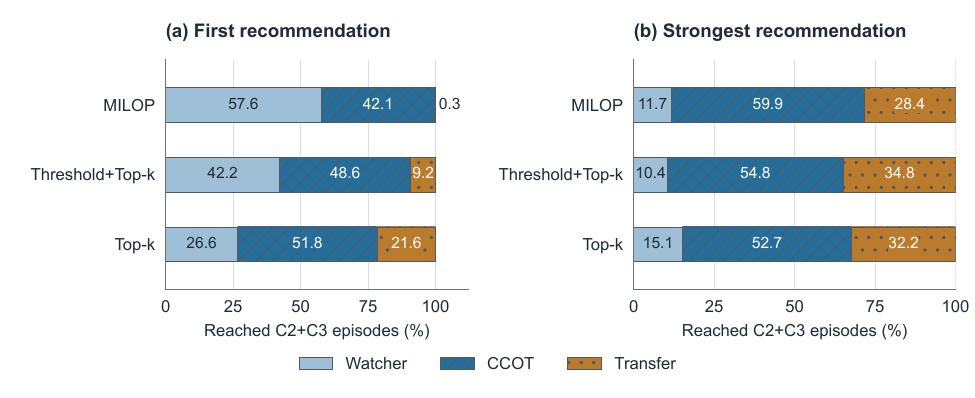}}
{First and strongest recommendations among reached severe episodes\label{fig:msom-uniform-S5-first-strongest-recommendation}}
{Policy-specific denominators are 726, 661, and 583 reached severe episodes for MILOP, Threshold+Top-$k$, and Top-$k$. Panels show first and strongest pre-departure actions as separate distributions, not matched transitions.}
\end{figure}

\subsection{Strongest-Action Attribution across Risk Bins}
\label{sec:msom-ec-bin-details}


For the all-recorded-start comparison in Figure~\ref{fig:msom-S7-action-start-bin}, the Watcher denominators are 2,276, 1,601, and 1,035 starts under MILOP, Threshold+Top-$k$, and Top-$k$; the CCOT denominators are 2,174, 2,242, and 1,531. MILOP initiates 92.4\% of Watcher starts in $B_3$ and 76.9\% of CCOT starts in $B_4$. Watcher starts instead concentrate in $B_2$ under both baselines. These percentages exclude non-ICU departures and are not conditional action probabilities among all decision opportunities.

Episode-level attribution credits each episode once to its strongest strictly pre-transfer action. The all-recorded-start distribution in Figure~\ref{fig:msom-S7-action-start-bin} includes repeated starts and applies no pre-exit restriction. Both figures pool $C_0$--$C_3$ episodes.

\begin{figure}[!htbp]
\FIGURE
{\includegraphics[width=\textwidth]{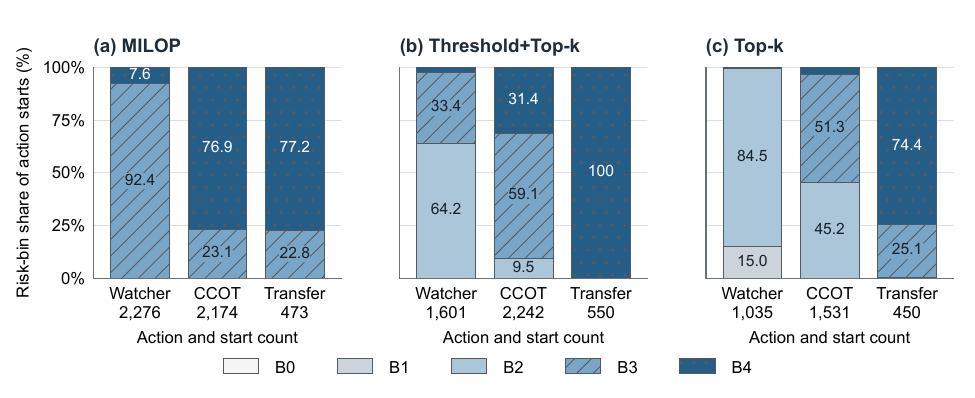}}
{Risk bin at recommendation initiation\label{fig:msom-S7-action-start-bin}}
{Within-action risk-bin shares among recorded starts in ICU-transfer episodes. Repeated starts are included without a pre-departure restriction; continuing assignments and non-ICU episodes are excluded. Counts below bars are starts, not episodes.}
\end{figure}

\subsection{Risk Profiles and Observed History among Missed Episodes}
\label{sec:msom-ec-miss-details}

Figure~\ref{fig:msom-S8-missed-final-bin} separates the final-bin distribution of misses by historical severity. MILOP's 68 critical misses comprise 7, 17, 34, and 10 episodes ending in $B_0$ through $B_3$, respectively, with none in $B_4$. Its 37 emergent misses comprise 6, 7, 18, and 6 in those bins, again with none in $B_4$. By contrast, Threshold+Top-$k$ leaves 49 critical and 9 emergent episodes unreached at a final $B_4$ observation; Top-$k$ leaves 118 and 15. Thus, the contrast at the highest final bin occurs in both severity groups, while most of MILOP's residual misses end in lower bins.

\begin{figure}[!htbp]
\FIGURE
{\includegraphics[width=\textwidth]{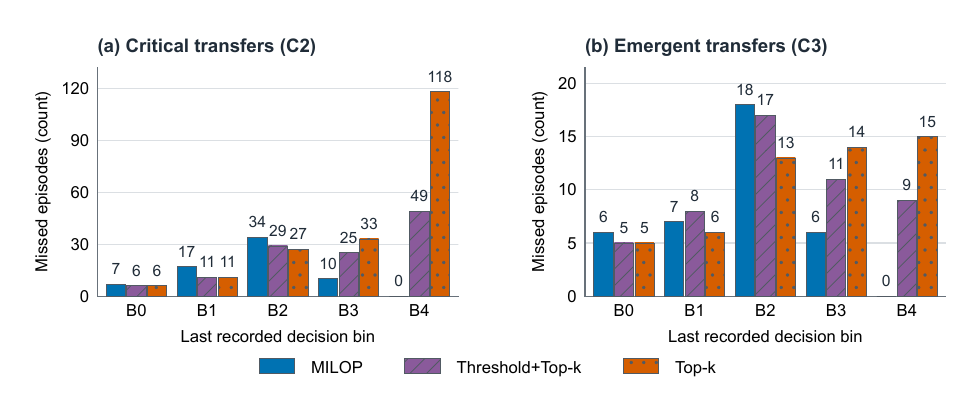}}
{Final recorded risk bin among missed critical and emergent episodes\label{fig:msom-S8-missed-final-bin}}
{Misses receive no qualifying recommendation strictly before transfer and are grouped by their last pre-transfer bin. Critical miss totals are 68/120/195 and emergent totals 37/50/53 for MILOP/Threshold+Top-$k$/Top-$k$. Panels use different scales and policy-specific missed populations.}
\end{figure}

Each missed episode is assigned to its last recorded pre-transfer decision bin. The figure compares counts among each policy's own missed episodes, not coverage within a common risk stratum or matched sets of missed patients. The final bin can differ from the highest observed bin: an episode can enter $B_3$ or $B_4$ earlier and finish in a lower bin. Among MILOP's critical and emergent misses, 80.9\% and 81.1\%, respectively, never enter either upper bin; the corresponding percentages are 34.2\% and 56.0\% under Threshold+Top-$k$ and 18.5\% and 43.4\% under Top-$k$.

Table~\ref{tab:msom-miss-diagnostics} summarizes observed history among each policy's missed severe episodes. History is measured from the number of recorded decision snapshots expressed in hours, rather than assumed to equal continuous ward length of stay. Among missed critical and emergent episodes, the respective medians are 25.5 and 13.0 hours for MILOP, 11.2 and 10.2 for Threshold+Top-$k$, and 9.5 and 10.0 for Top-$k$.

\begin{table}[H]
\TABLE
{Additional characteristics of missed severe episodes\label{tab:msom-miss-diagnostics}}
{\begin{tabular*}{\textwidth}{@{\extracolsep{\fill}}lrrr@{}}
\hline\up
Characteristic & MILOP & Threshold+Top-$k$ & Top-$k$ \down\\
\hline\up
Some monitoring allowance available & 98.1\% & 7.1\% & 0.0\% \\
Fewer than two hours of observed history & 6.7\% & 11.8\% & 10.5\% \\
Median observed history (hours) & 23.5 & 10.8 & 9.8 \down\\
\hline
\end{tabular*}}
{Each column describes that policy's own missed severe episodes: 105 under MILOP, 170 under Threshold+Top-$k$, and 248 under Top-$k$. Flags overlap. Monitoring availability means some Watcher or CCOT allowance remains at an observed pre-transfer snapshot; it need not coincide with a useful opportunity to act. Observed history is measured from recorded decision snapshots. These diagnostics do not identify causes of misses.}
\end{table}

\subsection{Lead-Time Definitions and Conditional Comparisons}
\label{sec:msom-ec-timing-details}

The warning-deadline comparison in the main manuscript retains all 831 severe episodes, including misses. Counts sum disjoint lead-time buckets at or above each boundary without interpolation; lower bounds are inclusive and the final bucket requires strictly positive lead. The paired comparison instead uses separate jointly reached cohorts, 641 episodes against Threshold+Top-$k$ and 551 against Top-$k$, and reports the median within-episode difference in first-recommendation time. Its median need not equal a difference of marginal medians, and the reported interquartile ranges are not confidence intervals.

Median leads among reached episodes are 19.2, 21.4, and 29.8 hours for severe episodes in MILOP, Threshold+Top-$k$, and Top-$k$ order. These exclude misses, and because each policy reaches a different set of episodes they are not a like-for-like comparison: MILOP reaches 85 severe episodes that Threshold+Top-$k$ does not and 175 that Top-$k$ does not, against 20 and 32 in the other direction. Episodes reached only by the broader policy are disproportionately short-stay and enter its denominator with short leads, so wider coverage lowers this statistic. Within the paired cohorts, where that effect is removed, MILOP's median lead is 23.6 and 31.1 hours against comparator values of 21.4 and 30.0.

\end{document}